\documentclass[lettersize,journal]{IEEEtran}
\usepackage[colorlinks,urlcolor=blue,linkcolor=blue,citecolor=blue]{hyperref}
\usepackage{amsmath,amsfonts}
\usepackage{algorithmicx}
\usepackage{algpseudocode}
\usepackage{titlesec} \titlespacing*{\paragraph}{0pt}{3pt}{0.5em}
\usepackage{array}
\usepackage[caption=false,font=normalsize,labelfont=sf,textfont=sf]{subfig}
\usepackage{textcomp}
\usepackage{stfloats}
\usepackage{url}
\usepackage{amsmath}
\usepackage[ruled,vlined,linesnumbered]{algorithm2e}
\usepackage[table]{xcolor}

\definecolor{semgreen}{RGB}{246,252,248}

\newcommand{\res}[2]{#1{\scriptsize$\pm$#2}}
\newcommand{\bestres}[2]{\textbf{#1}{\scriptsize$\pm$#2}}
\usepackage{graphicx}
\usepackage{verbatim}
\def\BibTeX{{\rm B\kern-.05em{\sc i\kern-.025em b}\kern-.08em
    T\kern-.1667em\lower.7ex\hbox{E}\kern-.125emX}}
\usepackage{balance}
\begin{document}
\title{Continual Visual Learning under Evolving Semantic Concept Shift}
\author{
Ismail Lamaakal, \IEEEmembership{Student Member, IEEE}, Chaymae Yahyati, \IEEEmembership{Student Member, IEEE}, {Yassine Maleh}, \IEEEmembership{Senior Member, IEEE}, Khalid El Makkaoui, \IEEEmembership{Senior Member, IEEE}, Ibrahim Ouahbi

\thanks{Lamaakal. I, Yahyati. C, El Makkaoui. K, Ouahbi. I are with the Department of Computer Science, Faculty of Applied Sciences Nador, Mohammed Premier University, Oujda, Morocco. e-mails: khalid.elmakkaoui@ieee.org, (chaymae.yahyati, i.ouahbi)@ump.ac.ma. and Corresponding author: (ismail.lamaakal@ieee.org)}
\thanks{Yassine Maleh and is with the Laboratory LaSTI, ENSAK, Sultan Moulay Slimane University, Khouribga, Morocco.
(e-mail: yassine.maleh@ieee.org).}

}

\maketitle


\begin{abstract}
Visual foundation models are commonly adapted under the assumption that
the appearance of incoming data may change while the semantic meaning
of the prediction task remains fixed. In long-lived visual systems,
however, taxonomies, policies, and concept definitions can themselves
evolve, causing the same visual evidence to require a different
interpretation. We study this setting as \emph{evolving semantic concept
shift} and introduce \textsc{SemReWrite}, a framework for selectively
updating obsolete visual--semantic mappings while preserving knowledge
that remains valid. SemReWrite represents changes between old and
revised semantic specifications, combines semantic discrepancy with
sparse revised supervision to localize affected visual regions, and
uses an input-dependent low-rank rewriting mechanism together with
structured semantic memory, preservation, and obsolete-decision
suppression. We further introduce \textsc{EvoShift-Bench}, spanning
ImageNet, iNaturalist, CUB-200-2011, and DomainNet, with semantic
transitions including class split, merge, boundary revision, insertion,
partial redefinition, recurrence, and mixed semantic--appearance shift.
To explicitly evaluate selective semantic revision, we introduce
Rewrite Accuracy (RA) and Preservation Accuracy (PA) for affected and
unaffected regions, respectively, Obsolete Retention (OR) for measuring
residual outdated semantic associations, and the \emph{Selective
Revision Score} (SRS), which jointly summarizes rewriting and
preservation performance. Experiments show that SemReWrite achieves a
stronger balance between learning revised semantics and retaining
unaffected knowledge than prompt replacement, conventional fine-tuning,
parameter-efficient adaptation, and continual-learning strategies.
The results highlight that successful adaptation under semantic
evolution requires not simply preventing forgetting, but selectively
preserving what remains true while rewriting what has become obsolete.
\end{abstract}




\section{Introduction}
\label{sec:introduction}

Large-scale visual foundation models have substantially changed how
recognition systems are developed and deployed. Models such as CLIP,
SigLIP, DINOv2, and recent vision--language architectures provide
transferable visual representations and enable recognition through
natural-language descriptions with little task-specific training
\cite{awais2025foundation,zhang2024vlm}. Their increasing use in
long-lived systems has motivated extensive research on domain
adaptation, test-time adaptation, parameter-efficient tuning, prompt
learning, and continual learning
\cite{liu2024fewshot,wang2024continual}. Most of these approaches,
however, are designed for situations in which the visual environment
may change while the semantic meaning of the prediction task remains
essentially fixed.

This assumption is appropriate for conventional distribution shift:
the same object may appear under a different camera, illumination,
weather condition, artistic style, corruption, or deployment domain
while retaining the same correct semantic interpretation. Domain and
test-time adaptation methods are primarily designed to recover
predictive performance under such changes in the observed target
distribution \cite{liang2025tta,wang2025otta}. Real systems, however,
can face a fundamentally different form of evolution. The observed
visual evidence may remain similar while the rule assigning meaning to
that evidence changes. Such changes are related to the broader notion
of concept drift and conditional concept shift, in which the
relationship between observations and predictions evolves
\cite{hinder2024drift,zhu2026concept}. We refer to the particular
setting studied here as \emph{evolving semantic concept shift}.
Unlike conventional adaptation, the objective is no longer simply to
recognize the same concepts under new appearances, but to update the
model when the semantic definitions themselves evolve.

Such revisions can arise naturally in long-lived decision systems. In
autonomous driving, an initial taxonomy may represent several road
vehicles by a single ``vehicle'' category, while a later operational
policy separates passenger, commercial, and emergency vehicles because
they require different downstream actions. In industrial inspection, a
component previously classified as defective according to one tolerance
may later be judged using a revised threshold together with an
additional deformation criterion. In ecological monitoring, taxonomic
revisions may split a previously recognized category into several
species or merge categories that were previously treated independently.
More generally, evolving environments require deployed learning systems
to continuously detect, interpret, and respond to changes in the
relationship between observed data and the required decision
\cite{hinder2024drift,wang2024continual}. In each of the examples above,
the visual evidence can remain unchanged while its correct semantic
interpretation evolves.

This setting creates a conflict that conventional adaptation methods are
not designed to resolve. Standard fine-tuning can learn a new decision
rule, but unrestricted parameter updates can interfere with knowledge
that should remain valid. Continual-learning approaches address the
opposite problem by explicitly combating catastrophic forgetting through
regularization, replay, parameter isolation, or other preservation
mechanisms \cite{wang2024continual}. Under semantic revision, however,
some previous knowledge has intentionally become incorrect and should
no longer be preserved. Changing only the language prompt provides
another possibility for vision--language models, and prompt-based
adaptation can efficiently modify downstream behavior
\cite{bulat2024lasp,liu2024fewshot}; nevertheless, a revised textual
description specifies what a concept should mean without necessarily
identifying which part of the visual decision space must change.
Successful adaptation therefore requires neither preserving everything
nor rewriting everything, but selectively modifying obsolete mappings
while protecting knowledge that remains semantically correct.

A further difficulty is that the visual stream itself may provide no
indication that such a change has occurred. If the same images are
interpreted according to a revised taxonomy, operational policy, or
semantic definition, an adaptation mechanism observing only unlabeled
images has no reliable evidence of the semantic revision when the
observable input distribution remains unchanged. This differs from
standard test-time adaptation, where adaptation is driven by target
observations drawn from a changed deployment distribution
\cite{liang2025tta,wang2025otta}. Recent work on concept-shift-aware
domain adaptation has also highlighted the importance of changes in the
conditional input--label relationship, rather than considering only
feature or label-distribution shift \cite{zhu2026concept}. Our setting
goes one step further by treating the semantic specification itself as
an explicit evolving object. Semantic evolution therefore requires side
information describing what changed and a small amount of revised
supervision that connects the new definition to the relevant visual
evidence.

We introduce \textsc{SemReWrite}, a framework for selectively rewriting
visual knowledge under evolving semantic concept shift. Given the old
and revised semantic specifications, the framework first represents the
semantic change between them and uses this information together with a
few revised examples to identify the visual regions likely to be
affected. An input-dependent rewriting mechanism then applies stronger
adaptation where the semantic mapping should change, while preservation
constraints protect unaffected knowledge. A structured semantic concept
memory maintains relationships between concepts across revisions and
supports operations such as splitting, merging, partial redefinition,
and recurrence.

The main contributions of this work are summarized as follows:
\begin{enumerate}

    \item We formulate \emph{evolving semantic concept shift} as a
    visual learning problem in which the semantic specification itself
    changes over time, distinguishing it from conventional covariate
    shift, domain adaptation, and class-incremental learning.

    \item We identify the fundamental limitation of adaptation based
    only on an unlabeled visual stream when semantic definitions change
    without an observable change in the input distribution. This
    motivates the use of explicit semantic side information together
    with sparse supervision under the revised specification.

    \item We propose \textsc{SemReWrite}, which combines semantic-change
    representation, affected-region localization, structured semantic
    memory, selective low-rank rewriting, preservation of still-valid
    knowledge, and suppression of obsolete semantic mappings.

    \item We introduce \textsc{EvoShift-Bench}, covering class split,
    merge, decision-boundary revision, insertion, partial redefinition,
    recurring semantic changes, and mixed semantic--appearance shifts,
    together with evaluation measures that separately quantify rewriting,
    preservation, and obsolete-rule retention.
\end{enumerate}

The remainder of this paper is organized as follows: 
Section~\ref{sec:related_work} reviews related work on domain and
test-time adaptation, concept drift, continual learning, vision--language
adaptation, and model editing.
Section~\ref{sec:methodology} introduces the proposed
\textsc{SemReWrite} framework, including the formulation of evolving
semantic concept shift, the identifiability analysis, semantic-change
representation, affected-concept localization, semantic concept memory,
selective concept rewriting, and the proposed selective-revision
evaluation measures.
Section~\ref{sec:experiments} presents \textsc{EvoShift-Bench},
implementation details, comparison methods, and extensive experiments
covering different semantic shift types, supervision budgets, mixed
semantic--appearance shifts, continual semantic evolution, ablation
studies, and failure-mode analysis.
Finally, Section~\ref{sec:conclusion} summarizes the main findings,
discusses the limitations of the current framework, and outlines
directions for future research.
\vspace{-0.3cm}
\section{Related Work}
\label{sec:related_work}

\textbf{Domain and test-time adaptation.}
Domain adaptation and test-time adaptation address performance
degradation caused by changes between training and deployment
distributions. Existing approaches use feature alignment, self-training,
distribution matching, semantic alignment, feature statistics, or
lightweight parameter updates to improve robustness when incoming data
differ from the training environment
\cite{li2024sfda,shi2024classimbalance}.
Recent vision--language adaptation further exploits the shared
image--text representation of pretrained multimodal models and adapts
prompts, contextual representations, or lightweight modules at test
time. For example, recent language-driven TTA methods adapt frozen VLMs
to target domains through parameter-efficient contextual modules
\cite{zhang2025cola}. These approaches substantially improve robustness
to deployment-time distribution changes, but their primary objective is
still to adapt a model to a changed data environment while retaining the
meaning of the recognition task. Our setting instead considers cases
where the correct interpretation of the visual evidence itself changes,
including situations in which its appearance may remain essentially
unchanged.

\textbf{Concept drift and conditional shift.}
Concept drift concerns temporal changes in the statistical relationship
governing prediction and has been studied extensively for non-stationary
data streams. Recent work examines concept-drift detection directly in
image streams and analyzes how different forms of gradual and abrupt
drift affect visual learning systems
\cite{tran2026imagedrift}. Other methods continuously adapt deep models
to evolving streams by maintaining multiple learners or selectively
updating models as the underlying data-generating process changes
\cite{chambers2025deepstream}. Active domain adaptation addresses a
related but different problem by selecting a limited number of
informative target samples for annotation and using them to facilitate
cross-domain adaptation \cite{tian2024uccda}. These approaches generally
treat drift or domain change as a property to be detected or compensated
for from evolving observations. Our problem is complementary: the
semantic specification is explicitly revised, and the central question
is how the model itself should be selectively rewritten when the meaning
assigned to previously learned labels changes. We therefore focus on
semantic-label evolution induced by revised taxonomies, policies, or
concept definitions rather than generic temporal non-stationarity.

\textbf{Continual learning and forgetting.}
Continual learning aims to acquire knowledge sequentially while limiting
catastrophic forgetting. Recent class-incremental learning research
organizes existing solutions around rehearsal, regularization,
parameter isolation, representation learning, and model expansion
\cite{zhou2024cilsurvey}. More recent approaches explicitly constrain
the optimization trajectory, for example through flatness-aware
optimization and orthogonal gradient projection, to reduce interference
with previously learned tasks \cite{yang2025flatness}. Continual
learning has also been extended to vision--language models, where frozen
image and text encoders together with expandable task-specific
projections can preserve previous multimodal knowledge
\cite{zhou2025vlmforgetting}. These mechanisms are closely related to
our setting because semantic specifications also evolve sequentially.
Their objective, however, is predominantly to prevent previous knowledge
from being lost. Under semantic revision, part of that previous
knowledge may have become explicitly incorrect. We therefore distinguish
harmful forgetting of still-valid concepts from desirable forgetting of
obsolete semantic associations rather than treating all forgetting as
equally undesirable.

\textbf{Vision--language adaptation and model editing.}
The language interface of modern vision--language models provides a
natural mechanism for expressing downstream tasks and revised concept
descriptions. Recent work adapts such models through test-time prompt
learning \cite{lu2025taskprompt} and prompt optimization designed to
retain generalization across classes and domains
\cite{zhang2025consistentprompt}. In parallel, model editing seeks to
modify specific knowledge in pretrained models while minimizing
unintended changes to unrelated behavior. Recent multimodal editing
methods have begun extending this principle to vision--language
architectures; for example, dynamic low-rank modules can encode and
activate individual multimodal edits without fully retraining the
backbone \cite{chen2025multimelo}. The broader model-editing literature
similarly emphasizes reliability, locality, generalization, and the
avoidance of collateral knowledge degradation
\cite{wang2024knowledgeediting}. Nevertheless, changing a textual prompt
or successfully editing an isolated fact does not guarantee that the
corresponding visual decision region has been modified correctly,
particularly when only part of an existing concept changes.
\textbf{\textsc{SemReWrite} therefore couples explicit semantic revision with
affected-region localization and selective visual rewriting, rather than
treating the revised textual description as an independent replacement
classifier.}
\vspace{-0.2cm}
\section{Methodology}
\label{sec:methodology}

In this section, we present \textsc{SemReWrite}, a selective concept rewriting framework designed for visual recognition systems whose semantic definitions evolve over time.

\begin{figure*}[htbp]
    \centering
    \includegraphics[width=\textwidth]{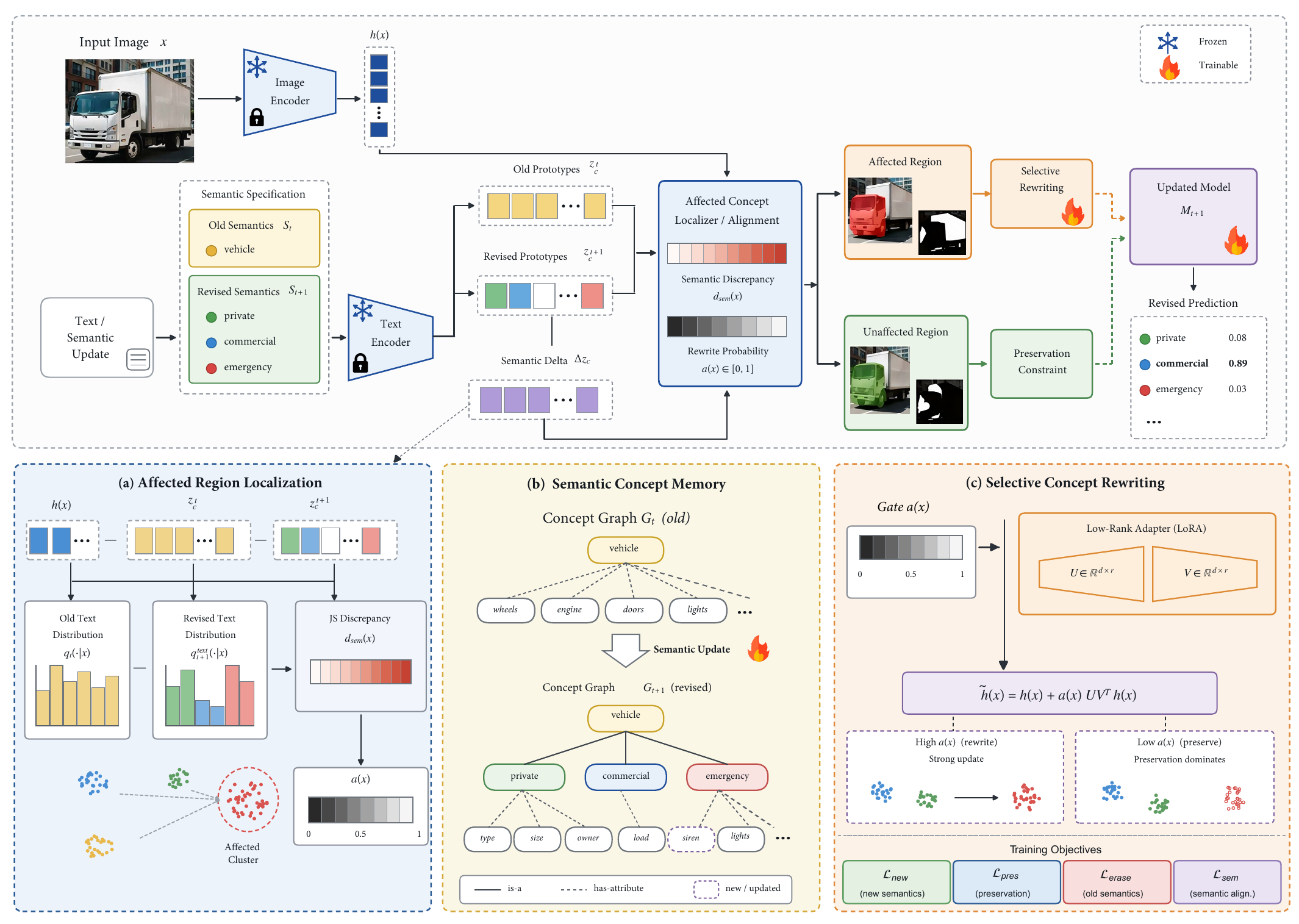}
    \caption{\textbf{\textit{Overview of the proposed \textsc{SemReWrite} framework.}}
    The old semantic specification $\mathcal{S}_{t}$ and the revised
    specification $\mathcal{S}_{t+1}$ are first encoded to characterize
    how the meaning of the task has changed. The resulting semantic
    change is incorporated into a structured semantic concept memory.
    In parallel, the frozen visual encoder maps each image into the
    shared representation space. The affected concept localizer combines
    the specification-level semantic discrepancy with sparse examples
    labeled under the new specification to estimate the rewrite
    probability $a(x)$. Inputs associated with a large rewrite
    probability activate the selective rewriting pathway, whereas inputs
    associated with stable semantics are constrained by the preservation
    pathway. The two branches jointly produce the updated model
    $M_{t+1}$, allowing obsolete semantic mappings to be replaced while
    retaining visual knowledge that remains valid.}
    \label{fig:semrewrite_framework}
\end{figure*}

As illustrated in Fig.~\ref{fig:semrewrite_framework}, the method first compares the old and revised semantic specifications, represents their difference in a shared vision-language space, identifies which concepts and visual samples are likely to be affected, updates a structured semantic memory, and finally performs selective parameter-efficient rewriting.


\vspace{-0.3cm}
\subsection{Problem Formulation}
\label{subsec:problem_formulation}

Let $\mathcal{X}$ denote the visual input space. At time $t$, the
recognition task is defined over
$\mathcal{Y}_t=\{1,\ldots,K_t\}$, where the number and identity of
concepts may change over time through insertion, removal, splitting, or
merging. Each concept $c\in\mathcal{Y}_t$ is associated with a semantic
description $s_c^t$, and the complete semantic specification is
$\mathcal{S}_t=\{s_c^t\}_{c\in\mathcal{Y}_t}$. Such descriptions may
encode natural-language definitions, attributes, hierarchical
relations, or task-specific decision criteria.

We factorize the data-generating process as
$P_t(X,Y\mid\mathcal{S}_t)=P_t(X)P_t(Y\mid X,\mathcal{S}_t)$.
Conventional covariate shift primarily changes $P_t(X)$ while retaining
the interpretation of the labels. In contrast, we define
\emph{semantic concept shift} as a revision
$\mathcal{S}_t\rightarrow\mathcal{S}_{t+1}$ for which
$P_t(Y\mid X,\mathcal{S}_t)\neq
P_{t+1}(Y\mid X,\mathcal{S}_{t+1})$ on a subset of non-zero probability
mass. Importantly, this can occur even when
$P_t(X)\approx P_{t+1}(X)$; the visual evidence may remain unchanged
while its correct semantic interpretation changes.

Because $\mathcal{Y}_t$ and $\mathcal{Y}_{t+1}$ need not coincide, we
compare the two semantic mechanisms on the common support
$\Omega_t=\mathcal{Y}_t\cup\mathcal{Y}_{t+1}$, assigning zero
probability to concepts absent at either time step. Let
$P_t^\star(\cdot\mid x)$ and $P_{t+1}^\star(\cdot\mid x)$ denote the
ground-truth old and revised conditional distributions on
$\Omega_t$. We define the affected region as
\begin{equation}
\mathcal{A}_t=
\left\{
x\in\mathcal{X}:
D_{\mathrm{TV}}
\left(
P_t^\star(\cdot\mid x),
P_{t+1}^\star(\cdot\mid x)
\right)>\delta_A
\right\},
\end{equation}
where
$D_{\mathrm{TV}}(p,q)=\frac{1}{2}\sum_{c\in\Omega_t}|p(c)-q(c)|$.
The complementary region
$\mathcal{U}_t=\mathcal{X}\setminus\mathcal{A}_t$ contains inputs whose
semantic interpretation remains valid. For deterministic labels,
$\delta_A=0$ reduces $\mathcal{A}_t$ to examples whose old and revised
assignments disagree.

The learner does not observe $\mathcal{A}_t$ or $\mathcal{U}_t$ during
adaptation; they are used only to formalize the problem and for
benchmark evaluation. Instead, it receives the previous model $M_t$,
the semantic transition $(\mathcal{S}_t,\mathcal{S}_{t+1})$, an
unlabeled pool
$\mathcal{D}_{t+1}^{u}=\{x_j\}_{j=1}^{N_u}$, and a small revised
supervision set
$\mathcal{D}_{t+1}^{\mathrm{few}}
=\{(x_i,y_i^{t+1})\}_{i=1}^{B}$ with $B\ll N_u$.

The objective is to obtain $M_{t+1}$ that implements the revised
semantic rule on $\mathcal{A}_t$ while preserving still-valid knowledge
on $\mathcal{U}_t$. This selective requirement distinguishes semantic
concept rewriting from both conventional adaptation, which primarily
addresses changes in the input distribution, and continual learning,
which generally seeks to preserve previously acquired knowledge.


\subsection{Identifiability of Semantic Concept Shift}
\label{subsec:identifiability}

A particularly important property of semantic concept shift is that it can be invisible in the unlabeled visual stream. This is fundamentally different from many domain shifts. For instance, if photographs are replaced by sketches, a change in $P(X)$ can often be detected using only the visual observations. However, if the same images should simply be interpreted using a revised semantic rule, the marginal image distribution may remain unchanged. Consider two environments, denoted by $H_0$ and $H_1$. Assume that both environments generate exactly the same visual marginal distribution, $P_{H_0}(X)=P_{H_1}(X)=P_X$, but use different conditional labeling mechanisms, so that $P_{H_0}(Y\mid X)\neq P_{H_1}(Y\mid X)$ over a subset of the input space. Suppose an observer receives only an unlabeled sample $X_{1:n}=\{X_1,\ldots,X_n\}$. Because every $X_i$ is drawn from the same $P_X$ under either environment, the complete observable likelihood is identical: $P(X_{1:n}\mid H_0)=P_X^{\otimes n}=P(X_{1:n}\mid H_1)$.

\textbf{Proposition 1.}
\emph{If two environments have identical visual marginals but different conditional labeling mechanisms, then no statistical decision rule using only unlabeled observations from the visual marginal can identify which labeling mechanism is currently active. Under equal priors, the minimum achievable binary discrimination error is $1/2$.}

The proof follows directly from equality of the observable distributions. Let $\phi(X_{1:n})\in\{0,1\}$ be any statistical test attempting to identify the active environment. Since the distribution of $X_{1:n}$ is identical under $H_0$ and $H_1$, the probability of every event produced by $\phi$ is also identical. Equivalently, the total-variation distance between the two observable distributions is zero. Therefore, no test based exclusively on the unlabeled visual sample has discriminatory information about the semantic change. A complete measure-theoretic proof is provided in the supplementary material.

This result has a direct methodological consequence. \textsc{SemReWrite} should not attempt to infer a purely semantic revision from unlabeled images alone because, in the general case, the required information is not present in those observations. Instead, we explicitly provide two complementary information sources. The first is the revised semantic specification $\mathcal{S}_{t+1}$, which communicates \emph{what the concepts now mean}. The second is the small set $\mathcal{D}_{t+1}^{\mathrm{few}}$, which provides evidence about \emph{where the revised semantic distinctions occur in the visual representation space}. 

The separation between ``what changed'' and ``where it changed'' is important. A natural-language instruction can state that a category has been divided into two subclasses, but it may not precisely specify the visual boundary separating them. Conversely, a few labeled images can indicate the revised boundary locally, but without an explicit semantic specification they provide only sparse information about the meaning and relationships of the new concepts. \textsc{SemReWrite} therefore combines both information sources rather than relying on either one independently.

\vspace{-0.2cm}
\subsection{Semantic Change Representation}
\label{subsec:semantic_representation}

The first operational component of \textsc{SemReWrite} converts the symbolic transition $\mathcal{S}_t\rightarrow\mathcal{S}_{t+1}$ into a continuous representation that can be compared with visual features. We employ the text encoder of a pretrained vision-language model for this purpose. Let $E_T(\cdot)$ denote the frozen text encoder and let its output dimension be $d$. Freezing the encoder is intentional: the shared pretrained representation provides a stable coordinate system in which the old and revised semantic definitions can be compared. If the text encoder itself were extensively adapted during every semantic transition, movement of the representation space could become confounded with movement of the concepts.

For each concept $c$ present at time $t$, the old semantic prototype is computed as $z_c^t=\operatorname{norm}(E_T(s_c^t))$, where $\operatorname{norm}(v)=v/\|v\|_2$ denotes $\ell_2$ normalization. Likewise, for a concept appearing in the revised specification, we compute $z_c^{t+1}=\operatorname{norm}(E_T(s_c^{t+1}))$. Normalization places the prototypes on the unit hypersphere and allows their compatibility with visual representations to be expressed using cosine similarity. When a concept exists in both specifications, its semantic displacement is represented by $\Delta z_c=z_c^{t+1}-z_c^t$. The vector $\Delta z_c$ should be interpreted as the direction in the language embedding space along which the definition of concept $c$ has moved. It does not by itself specify how model parameters should change; rather, it is an explicit representation of the semantic revision that will later guide localization and rewriting.

Label-space changes require slightly different treatment. If $c$ is newly introduced at time $t+1$, there is no old prototype to subtract, and we define its semantic change representation as $\Delta z_c=z_c^{t+1}$. If a concept disappears from the revised specification, its transition is represented by $\Delta z_c=-z_c^t$. Therefore, persistent modification, insertion, and deletion can all be represented within a common semantic-transition mechanism. For persistent concepts, we additionally quantify the magnitude of semantic movement using the cosine distance $d_c^{\mathrm{text}}=1-(z_c^t)^\top z_c^{t+1}$. Because both prototypes are normalized, their inner product is the cosine similarity. A value close to zero indicates that the old and revised descriptions occupy nearly the same direction in the pretrained language space, whereas a larger value indicates a more substantial semantic revision. This text-space difference is useful but should not be interpreted as the final estimate of concept shift. Language encoders capture semantic relationships at a high level, while the actual decision boundary is determined jointly by language and visual evidence. For example, two textual definitions may differ significantly while affecting only a small subset of images. Conversely, a seemingly modest change to a threshold-oriented definition may cause a large number of visual samples to switch labels. Consequently, \textsc{SemReWrite} uses the semantic representation to identify candidate changes and subsequently localizes their effect in the visual space.


\subsection{Affected Concept Localization}
\label{subsec:affected_localization}

The affected concept localizer estimates, for each input $x$, how
strongly the revised semantic specification should modify its
representation. As shown in Fig.~\ref{fig:semrewrite_framework}, this
score later controls the selective rewriting mechanism. Let $E_V$ be
the frozen visual encoder and
$h(x)=\operatorname{norm}(E_V(x))$ the normalized visual
representation. Using the old semantic prototypes $\{z_c^t\}$, we
compute
$q_t(c\mid x)=
\exp(h(x)^\top z_c^t/\tau)/
\sum_{k\in\mathcal{Y}_t}\exp(h(x)^\top z_k^t/\tau)$,
where $\tau>0$ is the temperature. Replacing the old prototypes with
the revised ones gives the text-induced distribution
$q_{t+1}^{\mathrm{text}}(c\mid x)=
\exp(h(x)^\top z_c^{t+1}/\tau)/
\sum_{k\in\mathcal{Y}_{t+1}}
\exp(h(x)^\top z_k^{t+1}/\tau)$.
No model parameters are updated at this stage; the two distributions
differ only through the semantic specification.

Because the label spaces may differ, both distributions are extended
to the common support $\Omega_t$ and denoted by
$\bar q_t$ and $\bar q_{t+1}^{\mathrm{text}}$. Their semantic
discrepancy is measured as
$d_{\mathrm{sem}}(x)=
\operatorname{JS}(\bar q_t(\cdot\mid x),
\bar q_{t+1}^{\mathrm{text}}(\cdot\mid x))/\log 2$,
where
$\operatorname{JS}(p,q)=
\frac{1}{2}D_{\mathrm{KL}}(p\|m)+
\frac{1}{2}D_{\mathrm{KL}}(q\|m)$ and $m=(p+q)/2$.
The normalization gives $d_{\mathrm{sem}}(x)\in[0,1]$.
A large value indicates that the revised semantic descriptions
substantially change the interpretation of $x$. We use JS divergence
because it is symmetric and remains finite when new or removed
categories receive zero probability under one specification.

Semantic discrepancy alone may overestimate the affected region because
changing a prototype can alter softmax probabilities even for examples
whose correct label should remain unchanged. We therefore complement it
with a lightweight sample-level localizer
$g_\psi(h(x))\in(0,1)$ trained from the sparse revised supervision.
Let $\mathcal{C}_t^{\mathrm{st}}$ denote the concepts whose semantics
remain unchanged, and let $\mathcal{C}_t(y)$ contain the stable old
concepts semantically equivalent to revised label $y$. For a revised
example $(x_i,y_i^{t+1})$, we define the soft change target
$r_i=1-\sum_{c\in\mathcal{C}_t(y_i^{t+1})}q_t(c\mid x_i)$.
Thus, $r_i$ is small when the old model already supports an unchanged
semantic interpretation and approaches one when the revised label has
no stable predecessor or conflicts with the old prediction.

The localizer is optimized using soft binary cross-entropy,
$\mathcal{L}_{\mathrm{gate}}
=-\frac{1}{B}\sum_{i=1}^{B}
[r_i\log g_\psi(h_i)+(1-r_i)\log(1-g_\psi(h_i))]$,
where $h_i=h(x_i)$. Finally, the specification-level and visual
localization signals are combined into the continuous rewrite gate
$a(x)=\eta d_{\mathrm{sem}}(x)+(1-\eta)g_\psi(h(x))$,
with $0\leq\eta\leq1$. Since both terms lie in $[0,1]$, we also have
$a(x)\in[0,1]$. The gate $a(x)$ is used directly during selective rewriting rather than
only for analysis: affected examples receive stronger representation
updates, while inputs predicted to remain semantically valid are
modified only weakly. This input-dependent control allows
\textsc{SemReWrite} to localize semantic revision instead of applying
the same update to the entire visual space.

\vspace{-0.3cm}
\subsection{Semantic Concept Memory}
\label{subsec:semantic_memory}

Semantic revisions often modify relations among concepts rather than
individual classes independently. To represent such structure,
\textsc{SemReWrite} maintains a typed semantic graph
$\mathcal{G}_t=(\mathcal{V}_t,\mathcal{E}_t,\rho_t)$, where
$\mathcal{V}_t$ contains class, attribute, part, parent, and child nodes,
$\mathcal{E}_t$ contains their relations, and $\rho_t$ specifies relation
types such as \emph{is-a}, \emph{part-of}, \emph{has-attribute}, and
\emph{subclass-of}. Semantic operations including split, merge,
insertion, and redefinition are therefore represented structurally as
$\mathcal{G}_t\rightarrow\mathcal{G}_{t+1}$. For example, a split adds
new child nodes beneath an existing parent, whereas a merge associates
multiple previously distinct concepts with a common semantic parent.

Each revised node $v$ is initialized by its normalized textual
representation $z_v^{t+1}$. For every class node $c$, we introduce a
small learnable semantic residual $\delta_c$ and define the adapted
prototype as
$\widetilde z_c^{t+1}=
\operatorname{norm}(z_c^{t+1}+\delta_c)$.
Thus, the revised text embedding remains the semantic anchor while
$\delta_c$ provides the limited visual correction required by the
few-shot revised examples. Non-class nodes remain fixed at their
text-derived representations unless explicitly modified by the semantic
transition. To prevent few-shot adaptation from distorting either individual
concept meanings or their graph structure, we impose a semantic
consistency loss composed of prototype anchoring and relational
preservation:
$\mathcal{L}_{\mathrm{sem}}=
\frac{1}{K_{t+1}}\sum_{c\in\mathcal{Y}_{t+1}}
[1-\cos(\widetilde z_c^{t+1},z_c^{t+1})]
+\frac{1}{|\mathcal{E}_{t+1}|}
\sum_{(u,v)\in\mathcal{E}_{t+1}}
[\cos(\widetilde z_u^{t+1},\widetilde z_v^{t+1})
-\cos(z_u^{t+1},z_v^{t+1})]^2$.
The first term keeps adapted class prototypes close to their revised
semantic definitions, while the second preserves pairwise relations
specified by $\mathcal{G}_{t+1}$. Together, they reduce semantic drift
and interference among related concepts during selective rewriting.


\subsection{Selective Concept Rewriting}
\label{subsec:selective_rewriting}

After identifying what changed and estimating where the revision applies,
we modify the visual representation through an input-dependent low-rank
update. Rather than fine-tuning the full backbone on the small revised
set, \textsc{SemReWrite} freezes the pretrained visual and text encoders
and introduces trainable matrices $U,V\in\mathbb{R}^{d\times r}$ with
$r\ll d$. Their product $UV^\top$ defines a compact residual
transformation with rank at most $r$.

Crucially, the residual is modulated by the localization score
$a(x)$ from Section~\ref{subsec:affected_localization}. The rewritten
representation is
$\widetilde h(x)=
\operatorname{norm}\!\left(
h(x)+a(x)UV^\top h(x)
\right)$.
Thus, when $a(x)\approx0$, the model remains close to the frozen
representation, whereas $a(x)\approx1$ activates the semantic rewrite.
This directly implements selective adaptation over the affected and
unaffected regions.

The adapted representation is compared with the revised semantic
prototype $\widetilde z_c^{t+1}$ through
$\ell_c(x)=\widetilde h(x)^\top\widetilde z_c^{t+1}/\tau$,
and the revised predictive distribution is
$p_{\theta'}(c\mid x,\mathcal{S}_{t+1})=
\exp(\ell_c(x))/
\sum_{k\in\mathcal{Y}_{t+1}}\exp(\ell_k(x))$.
The trainable parameters are
$\theta'=\{U,V,\{\delta_c\},\psi\}$, comprising the low-rank visual
rewrite, semantic residuals, and localization head; the foundation
encoders remain frozen.

\subsubsection{Learning the Revised Decision Rule}

The first objective teaches the model to predict according to the new semantic specification. For the revised labeled set $\mathcal{D}_{t+1}^{\mathrm{few}}$, we use the standard cross-entropy objective

\begin{equation}
\mathcal{L}_{\mathrm{new}}
=
-\frac{1}{B}
\sum_{i=1}^{B}
\log
p_{\theta'}
\left(
y_i^{t+1}\mid
x_i,\mathcal{S}_{t+1}
\right).
\end{equation}
Although this loss is conventional, its role in our framework is precise. The semantic specification communicates the intended meaning of the revised concepts, but the few labeled examples show how those definitions correspond to actual visual observations. Consequently, $\mathcal{L}_{\mathrm{new}}$ provides the direct supervision required to move the affected decision regions toward the new rule. Optimizing only $\mathcal{L}_{\mathrm{new}}$, however, would reduce the method to ordinary few-shot adaptation. It provides no explicit mechanism for retaining the parts of the previous model that remain correct. The remaining objectives therefore regulate \emph{where} learning should occur and \emph{what} knowledge should be retained or removed.

\subsubsection{Preserving Knowledge that Remains Valid}

Conventional distillation may be counterproductive under semantic
revision because matching the complete output of $M_t$ would also
preserve predictions that have become obsolete. We therefore distill
only concepts whose definitions remain unchanged and weight
preservation according to the estimated unaffectedness of each input.

Let $\mathcal{C}_t^{\mathrm{st}}$ denote the persistent concepts with
unchanged semantics, and let
$\mathcal{R}_{\mathcal{C}_t^{\mathrm{st}}}[p]$ denote restriction of
distribution $p$ to this set followed by renormalization. For the
unlabeled pool, the preservation objective is
$\mathcal{L}_{\mathrm{pres}}=
\frac{\sum_{x\in\mathcal{D}_{t+1}^{u}}(1-a(x))
D_{\mathrm{KL}}\!\left(
\mathcal{R}_{\mathcal{C}_t^{\mathrm{st}}}
[p_{\theta_t}(\cdot\mid x)]
\|
\mathcal{R}_{\mathcal{C}_t^{\mathrm{st}}}
[p_{\theta'}(\cdot\mid x)]
\right)}
{\sum_{x\in\mathcal{D}_{t+1}^{u}}(1-a(x))+\epsilon}$,
where $\epsilon>0$ ensures numerical stability. If no stable concept
exists, this term is omitted.

Restricting distillation to $\mathcal{C}_t^{\mathrm{st}}$ prevents the
new model from being explicitly constrained toward obsolete semantic
decisions, while the factor $1-a(x)$ concentrates preservation on
inputs predicted to remain unaffected. Thus, examples with
$a(x)\approx1$ are free to change, whereas those with $a(x)\approx0$
remain close to the previous model on still-valid concepts. Unlike
standard forgetting prevention, the objective is therefore to preserve
\emph{only knowledge that remains semantically valid}.

\subsubsection{Suppressing Obsolete Semantic Decisions}

Cross-entropy with the new label encourages the correct revised class to gain probability, but it does not explicitly measure whether the old, conflicting decision has been removed. This distinction is important for our problem because successful semantic adaptation requires both acquisition of the new relation and elimination of the outdated relation.

For a revised example $x_i$, let $\widehat y_i^t=\arg\max_{c\in\mathcal{Y}_t}q_t(c\mid x_i)$ denote the class favored by the old semantic model. We apply the erasure objective only when three conditions hold: the localizer assigns the sample a sufficiently high affected probability, the old predicted class still exists in the revised label space, and the old prediction disagrees with the revised ground-truth label.

Let $\mu_i$ be one when these conditions are satisfied and zero otherwise. The obsolete-decision objective is

\begin{equation}
\mathcal{L}_{\mathrm{erase}}
=
-
\frac{
\sum_{i=1}^{B}
\mu_i
\log
\left[
1-
p_{\theta'}
\left(
\widehat y_i^t\mid
x_i,\mathcal{S}_{t+1}
\right)
+\epsilon
\right]
}{
\sum_{i=1}^{B}\mu_i+\epsilon
}.
\end{equation}
Minimizing this term explicitly decreases the probability assigned to a conflicting old decision. We do not apply the term when the old class has been removed completely from $\mathcal{Y}_{t+1}$ because such a class no longer exists in the revised classifier and is therefore suppressed structurally. The distinction between $\mathcal{L}_{\mathrm{new}}$ and $\mathcal{L}_{\mathrm{erase}}$ will also be evaluated experimentally. The former asks whether the new semantic mapping can be learned, while the latter asks whether obsolete semantic evidence remains active after adaptation. Removing either component produces a different failure mode and therefore forms part of the ablation analysis in Section~\ref{sec:experiments}.

\subsubsection{Controlling the Magnitude of the Rewrite}

Because the revised supervision is intentionally sparse, the adaptation parameters should remain compact unless the data provide strong evidence for a larger modification. We therefore regularize the trainable residuals using $\mathcal{L}_{\mathrm{reg}}=\|U\|_F^2+\|V\|_F^2+\frac{1}{K_{t+1}}\sum_{c\in\mathcal{Y}_{t+1}}\|r_c\|_2^2$. The Frobenius penalties constrain the visual low-rank update, while the final term discourages semantic prototypes from drifting unnecessarily far from their text-derived initialization.

The complete learning objective combines the revised supervision, preservation, obsolete-decision suppression, semantic consistency, gate supervision, and residual regularization:

\begin{equation}
\mathcal{L}_{\mathrm{total}}
=
\mathcal{L}_{\mathrm{new}}
+
\lambda_p\mathcal{L}_{\mathrm{pres}}
+
\lambda_e\mathcal{L}_{\mathrm{erase}}
+
\lambda_s\mathcal{L}_{\mathrm{sem}}
+
\lambda_g\mathcal{L}_{\mathrm{gate}}
+
\lambda_r\mathcal{L}_{\mathrm{reg}}.
\end{equation}
The coefficients $\lambda_p$, $\lambda_e$, $\lambda_s$, $\lambda_g$, and $\lambda_r$ are non-negative hyperparameters controlling the relative importance of each constraint. Their interpretation is direct. Increasing $\lambda_p$ makes the model more conservative in stable regions; increasing $\lambda_e$ more strongly suppresses conflicting old decisions; $\lambda_s$ controls adherence to the revised semantic graph; $\lambda_g$ controls the supervision strength of the affected-region estimator; and $\lambda_r$ controls the overall complexity of the adaptation.

These objectives are complementary rather than redundant. $\mathcal{L}_{\mathrm{new}}$ determines what the updated prediction should be, $\mathcal{L}_{\mathrm{erase}}$ removes the specific relation that has become incorrect, and $\mathcal{L}_{\mathrm{pres}}$ protects relationships that should not change. Meanwhile, $\mathcal{L}_{\mathrm{sem}}$ constrains the semantic geometry and $\mathcal{L}_{\mathrm{gate}}$ determines where rewriting should be activated. Their combination implements the core principle of \textsc{SemReWrite}: \emph{preserve what remains true and rewrite what has become false}.

\vspace{-0.2cm}
\subsection{Measuring Selective Semantic Revision}
\label{subsec:selective_metrics}

Conventional classification accuracy measures whether the final prediction is correct, but it does not reveal \emph{why} adaptation succeeded or failed. This distinction is especially important in our setting. A model can improve its accuracy on revised concepts while unnecessarily damaging stable concepts. Conversely, a strongly regularized continual learner can preserve old performance while failing to replace a decision rule that has become obsolete. Evaluating only aggregate accuracy would hide both behaviors.

We therefore separately evaluate the two regions introduced in Section~\ref{subsec:problem_formulation}. The affected evaluation set is denoted by $\mathcal{D}_{A}=\{(x_i,y_i^{t+1}):x_i\in\mathcal{A}_t\}$, while the unaffected evaluation set is $\mathcal{D}_{U}=\{(x_i,y_i^{t+1}):x_i\in\mathcal{U}_t\}$. These partitions are derived from the known benchmark construction and are used strictly for offline evaluation; the ground-truth affected-region indicator is never provided to \textsc{SemReWrite} during adaptation. For a revised prediction $\widehat y_i^{t+1}$, we define \emph{Rewrite Accuracy} (RA) as the fraction of affected examples correctly classified according to the new semantics,

\begin{equation}
\mathrm{RA}
=
\frac{1}{|\mathcal{D}_{A}|}
\sum_{(x_i,y_i)\in\mathcal{D}_{A}}
\mathbb{I}
\left[
\widehat y_i^{t+1}=y_i
\right].
\end{equation}
RA therefore measures the ability to replace an outdated semantic rule. A high RA indicates that the model has successfully modified its behavior where a modification was required.

We complement RA with \emph{Preservation Accuracy} (PA), defined on the unaffected region as

\begin{equation}
\mathrm{PA}
=
\frac{1}{|\mathcal{D}_{U}|}
\sum_{(x_i,y_i)\in\mathcal{D}_{U}}
\mathbb{I}
\left[
\widehat y_i^{t+1}=y_i
\right].
\end{equation}
PA measures whether the adaptation process has retained knowledge whose semantic validity has not changed. It therefore directly captures harmful forgetting in the stable portion of the task. Neither RA nor PA should be interpreted independently. A model that aggressively retrains on revised examples may achieve high RA but lose PA, whereas a highly conservative model may obtain high PA while retaining obsolete behavior and therefore produce a poor RA. We consequently summarize the balance between both objectives using the \emph{Selective Revision Score} (SRS),

\begin{equation}
\mathrm{SRS}
=
\frac{
2\,\mathrm{RA}\,\mathrm{PA}
}{
\mathrm{RA}+\mathrm{PA}
},
\end{equation}
with $\mathrm{SRS}=0$ when $\mathrm{RA}+\mathrm{PA}=0$. When RA and PA are reported as proportions, SRS lies in $[0,1]$.

We use a harmonic mean rather than an arithmetic mean because the harmonic mean penalizes strongly imbalanced behavior. For example, achieving nearly perfect preservation while completely failing to learn the revised concept should not be considered successful semantic adaptation. Similarly, perfectly fitting revised concepts while destroying previously correct stable concepts should also receive a low score. A high SRS can therefore be obtained only when the method jointly achieves successful semantic rewriting and strong knowledge preservation. In addition to RA, PA, and SRS, the experimental analysis will report the probability retained by obsolete concepts after adaptation. This additional measurement distinguishes merely predicting the new class correctly from genuinely suppressing the old semantic association. Together, these measurements enable the results in Section~\ref{sec:experiments} to separate desired semantic forgetting from catastrophic forgetting and to determine whether each component of \textsc{SemReWrite} performs its intended role.

\section{Experiments}
\label{sec:experiments}

We conduct extensive experiments to evaluate whether \textsc{SemReWrite}
can adapt to evolving semantic definitions while preserving knowledge
that remains valid. Our evaluation focuses on four questions: 
(i) whether the method generalizes across different forms of semantic
concept shift; (ii) whether selective rewriting provides a better
revision--preservation trade-off than conventional adaptation;
(iii) whether the approach remains effective with only a few examples
labeled according to the revised semantics; and (iv) whether its
behavior is consistent across different foundation-model backbones.
We therefore evaluate \textsc{SemReWrite} on the proposed
\textsc{EvoShift-Bench}, compare it with representative adaptation and
continual-learning approaches, and report both conventional accuracy
and the selective-revision metrics introduced in
Section~\ref{subsec:selective_metrics}.

\vspace{-0.3cm}
\subsection{EvoShift-Bench}
\label{subsec:evoshift_bench}

We construct \textsc{EvoShift-Bench} from four complementary visual
benchmark families: ImageNet, iNaturalist, CUB-200-2011, and DomainNet,
as summarized in Table~\ref{tab:benchmark_stats}. These datasets provide
complementary sources of semantic structure, including hierarchical
object relations, biological taxonomies, fine-grained visual
attributes, and cross-domain appearance variation
\cite{chatterjee2024imagenet,mason2025inaturalist,
ren2025cubattributes,li2024domainnet}. The benchmark is designed to
evaluate semantic evolution rather than arbitrary label permutation.
For an image with an immutable atomic annotation $u_i$, the semantic
specification at time $t$ determines its active label through
$y_i^t=g_t(u_i)$, whereas a revised specification produces
$y_i^{t+1}=g_{t+1}(u_i)$. The image itself therefore remains unchanged
while its semantic interpretation may change. Samples satisfying
$g_t(u_i)\neq g_{t+1}(u_i)$ constitute the ground-truth affected region
used only for evaluation; this information is never provided to the
adaptation method.

\begin{table}[htbp]
\centering
\caption{Overview of the principal \textsc{EvoShift-Bench} families.}
\label{tab:benchmark_stats}
\setlength{\tabcolsep}{5pt}
\renewcommand{\arraystretch}{1.08}
\footnotesize
\resizebox{\columnwidth}{!}{%
\begin{tabular}{l c c c c c c c}
\hline
\textbf{Dataset} &
\textbf{Classes} &
\textbf{Split} &
\textbf{Merge} &
\textbf{Boundary} &
\textbf{Insertion} &
\textbf{Recurring} &
\textbf{Mixed} \\
\hline
ImageNet     & 1,000 & \checkmark & \checkmark & --         & \checkmark & \checkmark & \checkmark \\
iNaturalist  & 1,000 & \checkmark & \checkmark & --         & \checkmark & \checkmark & -- \\
CUB-200      & 200   & --         & --         & \checkmark & --         & \checkmark & -- \\
DomainNet    & 345   & \checkmark & \checkmark & --         & \checkmark & --         & \checkmark \\
\hline
\end{tabular}}
\end{table}

\textbf{ImageNet.}
We use the WordNet hierarchy associated with ImageNet-1K to construct
semantically meaningful hierarchical transitions. ImageNet categories
are organized according to WordNet synsets and their hierarchical
relations, making the benchmark particularly suitable for reasoning
over coarse and fine semantic granularity
\cite{chatterjee2024imagenet}. Parent--child relations are used to
generate coarse-to-fine class splits, fine-to-coarse merges,
new-category insertion, and partial semantic redefinitions. For example,
a coarse concept can be replaced by multiple valid descendant concepts
under the revised specification. ImageNet therefore evaluates whether
the model can modify the granularity of previously learned object
concepts rather than simply recognizing entirely new visual categories.

\textbf{iNaturalist.}
We use the hierarchical taxonomy of iNaturalist to construct transitions
between family, genus, and species levels. iNaturalist observations are
organized through a continuously maintained biological taxonomy
covering multiple taxonomic ranks, and contemporary analyses emphasize
its importance for large-scale biodiversity recognition and research
\cite{mason2025inaturalist}. These natural taxonomic relations provide
class-split, class-merge, category-insertion, and recurring
semantic-shift scenarios. In particular, transitions such as
family$\rightarrow$genus$\rightarrow$family allow us to evaluate
whether a model can recover a semantic interpretation that becomes
valid again after an intermediate revision.

\textbf{CUB-200-2011.}
CUB is used to construct decision-boundary revisions that do not require
a change in the number of classes. The dataset contains 200 bird
categories and rich attribute-level supervision, including 312
attributes describing fine-grained visual characteristics
\cite{ren2025cubattributes}. These annotations allow semantic concepts
to be defined through interpretable properties such as wing color, bill
shape, breast pattern, or body appearance. The revised specification
changes which attributes determine the semantic decision rule while
keeping the underlying images fixed. This setting is important because
it distinguishes semantic revision from standard class-incremental
learning.

\textbf{DomainNet and ImageNet variants.}
Finally, we consider mixed semantic and appearance shifts. DomainNet is
widely used for evaluating transfer under substantial cross-domain
variation and contains visually different domains such as real images,
clipart, paintings, and sketches \cite{li2024domainnet}. Selected
ImageNet corruption variants provide an additional controlled source of
appearance shift and are widely used for evaluating adaptation under
changes such as noise and blur \cite{tang2024vct}. In these experiments,
the semantic specification and the visual distribution change
simultaneously, creating a more challenging scenario in which neither
ordinary domain adaptation nor semantic prompt replacement alone is
sufficient.

We organize the benchmark into seven transition types, summarized in
Table~\ref{tab:shift_protocols}. S1 splits an existing concept into
finer concepts, whereas S2 merges multiple concepts into a common
semantic category. S3 changes the decision boundary while preserving
the class vocabulary, and S4 introduces a new semantic category from a
previously broader concept. S5 modifies only part of the semantic
system, allowing selective preservation to be evaluated directly. S6
considers recurring specifications of the form
$\mathcal{S}_1\rightarrow\mathcal{S}_2\rightarrow\mathcal{S}_1$.
Finally, S7 combines semantic concept shift with a simultaneous
covariate shift. S3 and S5 are particularly important because they
change the underlying semantic decision rule without reducing the
problem to simple label-space expansion.

\begin{table}[htbp]
\centering
\caption{Semantic shift protocols in \textsc{EvoShift-Bench}.}
\label{tab:shift_protocols}
\setlength{\tabcolsep}{3pt}
\renewcommand{\arraystretch}{1.08}
\footnotesize
\begin{tabular}{c l l}
\hline
\textbf{ID} & \textbf{Shift} & \textbf{Example} \\
\hline
S1 & Split & Vehicle $\rightarrow$ Car / Truck \\
S2 & Merge & Husky + Malamute $\rightarrow$ Northern dog \\
S3 & Boundary revision & Semantic rule changes \\
S4 & Class insertion & Parent $\rightarrow$ new subclass \\
S5 & Partial redefinition & Subset of concepts changes \\
S6 & Recurring shift & $\mathcal{S}_1\!\rightarrow\!\mathcal{S}_2
                        \!\rightarrow\!\mathcal{S}_1$ \\
S7 & Mixed shift & Semantic + covariate shift \\
\hline
\end{tabular}
\end{table}

\vspace{-0.3cm}
\subsection{Implementation Details}
\label{subsec:implementation}

We use CLIP ViT-B/16 \cite{radford2021clip} as the primary backbone and
additionally evaluate SigLIP \cite{zhai2023siglip} and DINOv2
\cite{oquab2024dinov2} to examine backbone generalization. For DINOv2,
a lightweight projection connects its frozen visual representation to
the text-semantic space. Unless stated otherwise, the pretrained
foundation encoders remain frozen and only the localization head,
semantic residuals, and low-rank rewriting parameters are optimized.
The default low-rank dimension is $r=8$. We use AdamW
\cite{loshchilov2019adamw} for optimization and evaluate
revised-supervision budgets of $B\in\{1,2,4,8,16\}$ examples per
changed concept, with four examples used as the default setting. All
compared supervised methods receive the same revised examples for each
random seed. Experiments are repeated three times with independent
seeds, and results are reported as mean$\pm$standard deviation. In
addition to revised-task accuracy, we report RA, PA, SRS, and OR, where lower OR indicates more effective
removal of outdated semantic decisions.

\vspace{-0.2cm}
\subsection{Comparison Methods}
\label{subsec:baselines}

We compare \textsc{SemReWrite} with six groups of representative
baselines. \emph{Old Model} evaluates the previous model without any
adaptation, while \emph{New Text Only} replaces the old semantic
descriptions with the revised ones without learning; the latter directly
tests whether changing the language prompt alone is sufficient.
Standard adaptation baselines include full fine-tuning, classifier
fine-tuning, and linear probing. Parameter-efficient approaches include
LoRA \cite{hu2022lora}, lightweight adapters
\cite{houlsby2019adapters}, CoOp \cite{zhou2022coop}, and CoCoOp
\cite{zhou2022cocoop}. We additionally compare with continual-learning
methods including EWC \cite{kirkpatrick2017ewc}, LwF
\cite{li2018lwf}, Experience Replay \cite{rolnick2019replay}, and
DER++ \cite{buzzega2020der}, which are particularly relevant because
they explicitly aim to reduce catastrophic forgetting.

For vision--language adaptation, we include representative prompt- and
test-time adaptation approaches. TPT adapts textual prompts for each
test sample through entropy minimization \cite{shu2022tpt}, while
PromptAlign additionally aligns test-time feature statistics with the
source distribution \cite{samadh2023promptalign}. We further evaluate
TDA, a training-free dynamic adapter based on evolving key--value
caches \cite{karmanov2024tda}; StatA, which introduces a statistical
anchor for realistic test-time adaptation \cite{zanella2025stata}; and
BCA, which performs Bayesian adaptation of both class embeddings and
class priors \cite{zhou2025bca}, whenever their original protocols are
applicable to the considered setting. Finally, \emph{Oracle-Retrain} is
trained using the complete revised training set and serves only as a
fully supervised upper reference rather than as a budget-matched
competitor. All supervised budget-matched baselines use exactly the
same revised examples as \textsc{SemReWrite}.


\subsection{Main Results}
\label{subsec:main_results}

We first compare the methods across the principal semantic shift
settings under the default four-shot supervision budget.
Table~\ref{tab:main_results} reports conventional classification
accuracy together with the Selective Revision Score (SRS). The five
settings cover both label-structure changes, including split, merge,
and insertion, and genuine decision-rule changes, including boundary
revision and partial semantic redefinition.

\begin{table}[htbp]
\centering
\caption{Overall performance under five semantic shift types using four
revised examples per changed concept. Acc. measures revised-task
accuracy and SRS jointly measures rewriting and preservation.}
\label{tab:main_results}

\setlength{\tabcolsep}{4.2pt}
\renewcommand{\arraystretch}{1.10}
\footnotesize

\resizebox{\columnwidth}{!}{%
\begin{tabular}{l c c c c c c c}
\hline

\textbf{Method} &
\textbf{Metric} &
\textbf{Split} &
\textbf{Merge} &
\textbf{Boundary} &
\textbf{Redefinition} &
\textbf{Insertion} &
\textbf{Avg.}
\\
\hline

Old Model
& Acc.
& \res{65.2}{0.5}
& \res{69.1}{0.4}
& \res{61.0}{0.6}
& \res{63.1}{0.7}
& \res{58.5}{0.8}
& 63.4
\\
& SRS
& \res{65.2}{0.6}
& \res{68.1}{0.5}
& \res{62.0}{0.7}
& \res{63.4}{0.6}
& \res{59.0}{0.8}
& 63.5
\\
\hline

New Text
& Acc.
& \res{74.6}{0.6}
& \res{76.8}{0.5}
& \res{70.5}{0.8}
& \res{72.4}{0.7}
& \res{70.8}{0.7}
& 73.0
\\
& SRS
& \res{75.4}{0.6}
& \res{77.2}{0.5}
& \res{72.0}{0.7}
& \res{73.8}{0.7}
& \res{71.9}{0.8}
& 74.1
\\
\hline

Full FT
& Acc.
& \res{82.1}{0.7}
& \res{83.6}{0.6}
& \res{80.4}{0.8}
& \res{81.0}{0.7}
& \res{78.8}{0.9}
& 81.2
\\
& SRS
& \res{81.0}{0.7}
& \res{82.1}{0.6}
& \res{79.5}{0.8}
& \res{80.0}{0.7}
& \res{77.8}{0.9}
& 80.1
\\
\hline

LoRA
& Acc.
& \res{83.4}{0.5}
& \res{84.2}{0.5}
& \res{81.6}{0.7}
& \res{82.3}{0.6}
& \res{80.1}{0.8}
& 82.3
\\
& SRS
& \res{83.2}{0.5}
& \res{84.0}{0.5}
& \res{81.0}{0.7}
& \res{81.8}{0.6}
& \res{79.6}{0.8}
& 81.9
\\
\hline

LwF
& Acc.
& \res{79.8}{0.7}
& \res{81.4}{0.6}
& \res{76.6}{0.9}
& \res{78.1}{0.8}
& \res{75.2}{1.0}
& 78.2
\\
& SRS
& \res{81.8}{0.6}
& \res{83.0}{0.5}
& \res{78.9}{0.8}
& \res{79.6}{0.7}
& \res{76.8}{0.9}
& 80.0
\\
\hline

Replay
& Acc.
& \res{81.0}{0.6}
& \res{82.4}{0.6}
& \res{78.2}{0.8}
& \res{79.5}{0.7}
& \res{76.8}{0.9}
& 79.6
\\
& SRS
& \res{83.0}{0.6}
& \res{84.1}{0.5}
& \res{80.5}{0.7}
& \res{81.3}{0.7}
& \res{78.4}{0.8}
& 81.5
\\
\hline

StatA
& Acc.
& \res{78.3}{0.8}
& \res{80.0}{0.7}
& \res{74.1}{1.0}
& \res{76.0}{0.8}
& \res{73.5}{1.0}
& 76.4
\\
& SRS
& \res{79.4}{0.7}
& \res{80.8}{0.7}
& \res{76.5}{0.9}
& \res{77.8}{0.8}
& \res{75.0}{0.9}
& 77.9
\\
\hline

\rowcolor{semgreen}
\textbf{SemReWrite}
& Acc.
& \bestres{88.7}{0.4}
& \bestres{89.4}{0.4}
& \bestres{87.1}{0.5}
& \bestres{87.9}{0.4}
& \bestres{86.3}{0.6}
& \textbf{87.9}
\\

\rowcolor{semgreen}
& SRS
& \bestres{91.6}{0.3}
& \bestres{92.0}{0.3}
& \bestres{90.1}{0.4}
& \bestres{90.7}{0.4}
& \bestres{89.3}{0.5}
& \textbf{90.7}
\\
\hline

\end{tabular}}
\end{table}

Table~\ref{tab:main_results} shows that updating only the textual
specification already improves substantially over the unchanged model,
confirming that the new semantic description contains useful task
information. However, the remaining gap to learned adaptation is
considerable, particularly for boundary revision and partial
redefinition. These cases require the visual decision boundary itself
to change and therefore cannot be solved by simply replacing class
prompts. Full FT substantially improves the revised-task accuracy but provides a
weaker SRS, suggesting that its global update also modifies knowledge
whose semantics remain valid. LoRA improves this trade-off by
restricting parameter changes, whereas LwF and Replay preserve old
knowledge more strongly but adapt less effectively to revised semantic
rules. \textsc{SemReWrite} achieves the strongest performance across
all five shift types, including boundary revision and redefinition,
where the label space need not expand.


\subsubsection{Selective Forgetting}
\label{subsec:selective_forgetting}

To determine whether this improvement comes from modifying the correct
knowledge, Table~\ref{tab:forgetting_results} separately reports
RA, PA, OR, and SRS. RA evaluates samples whose semantic interpretation
changes, while PA evaluates samples that should remain unchanged.
Lower OR indicates more effective suppression of invalid old
predictions.

\begin{table}[htbp]
\centering
\caption{Revision--preservation analysis. RA evaluates changed regions,
PA evaluates stable regions, and OR measures retained obsolete
predictions.}
\label{tab:forgetting_results}

\setlength{\tabcolsep}{3.4pt}
\renewcommand{\arraystretch}{1.08}
\footnotesize

\begin{tabular}{l c c c c}
\hline
\textbf{Method}
&
\textbf{RA $\uparrow$}
&
\textbf{PA $\uparrow$}
&
\textbf{OR $\downarrow$}
&
\textbf{SRS $\uparrow$}
\\
\hline

Old Model
& \res{51.6}{0.7}
& \res{94.2}{0.3}
& \res{41.8}{1.0}
& \res{66.7}{0.6}
\\

New Text
& \res{65.8}{0.8}
& \res{92.5}{0.4}
& \res{29.7}{0.9}
& \res{76.9}{0.6}
\\

Full FT
& \res{84.7}{0.6}
& \res{80.1}{0.9}
& \res{13.6}{0.7}
& \res{82.3}{0.7}
\\

LoRA
& \res{83.6}{0.5}
& \res{84.9}{0.7}
& \res{14.8}{0.6}
& \res{84.2}{0.5}
\\

EWC
& \res{73.8}{0.7}
& \res{92.4}{0.4}
& \res{27.1}{0.8}
& \res{82.1}{0.5}
\\

LwF
& \res{76.2}{0.6}
& \res{92.1}{0.4}
& \res{24.6}{0.7}
& \res{83.4}{0.5}
\\

Replay
& \res{78.4}{0.7}
& \res{91.0}{0.5}
& \res{21.9}{0.7}
& \res{84.2}{0.6}
\\

StatA
& \res{72.8}{0.8}
& \res{90.2}{0.5}
& \res{26.8}{0.9}
& \res{80.6}{0.6}
\\
\hline

\rowcolor{semgreen}
\textbf{SemReWrite}
& \bestres{88.9}{0.4}
& \bestres{93.4}{0.3}
& \bestres{7.9}{0.4}
& \bestres{91.1}{0.3}
\\

\hline
\end{tabular}
\end{table}
The results expose two different failure modes. Full FT obtains high
RA and low OR but considerably reduces PA, indicating that global
adaptation successfully rewrites the changed concepts while also
damaging stable ones. Continual-learning methods show the opposite
behavior: EWC, LwF, and Replay retain high PA but also preserve a larger
fraction of obsolete decisions, which limits RA. \textsc{SemReWrite} combines the highest RA with a PA close to that of
the unchanged model and substantially reduces OR. The result supports
the distinction between catastrophic forgetting and desired semantic
forgetting: previous knowledge should be preserved only when its
semantic validity remains unchanged.

\begin{figure}[htbp]
    \centering
    \includegraphics[width=\columnwidth]
    {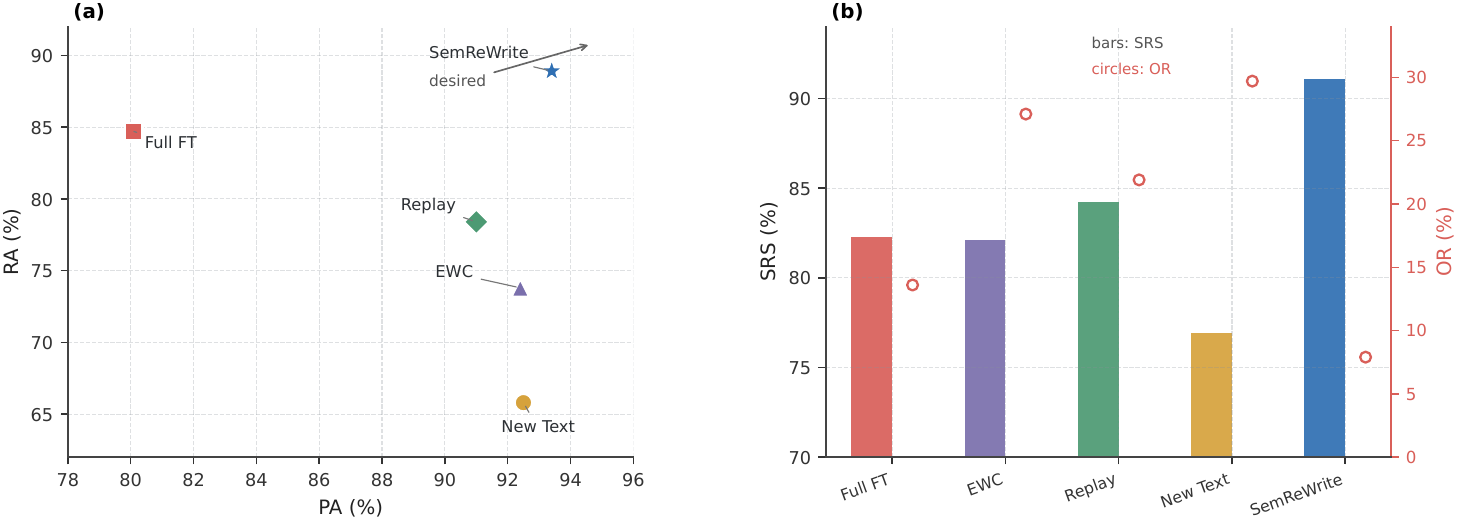}
    \caption{Selective forgetting analysis. (a) RA--PA trade-off;
    the desired region is the upper right. (b) SRS and obsolete
    retention (OR). SemReWrite combines stronger rewriting and
    preservation with lower obsolete retention.}
    \label{fig:selective_forgetting}
\end{figure}

Figure~\ref{fig:selective_forgetting}(a) makes the trade-off explicit.
Full FT is shifted toward strong rewriting but weak preservation,
whereas EWC and Replay favor preservation. SemReWrite occupies the
high-RA/high-PA region. Figure~\ref{fig:selective_forgetting}(b)
provides the complementary view: its high SRS is accompanied by the
lowest OR rather than by simply preserving previous predictions.


\subsubsection{Annotation Efficiency}
\label{subsec:annotation_efficiency}

We next vary the revised supervision budget from one to sixteen
examples per changed concept. Table~\ref{tab:annotation_efficiency}
reports RA, PA, and SRS for representative adaptation strategies.

\begin{figure}[htbp]
    \centering
    \includegraphics[width=\columnwidth]
    {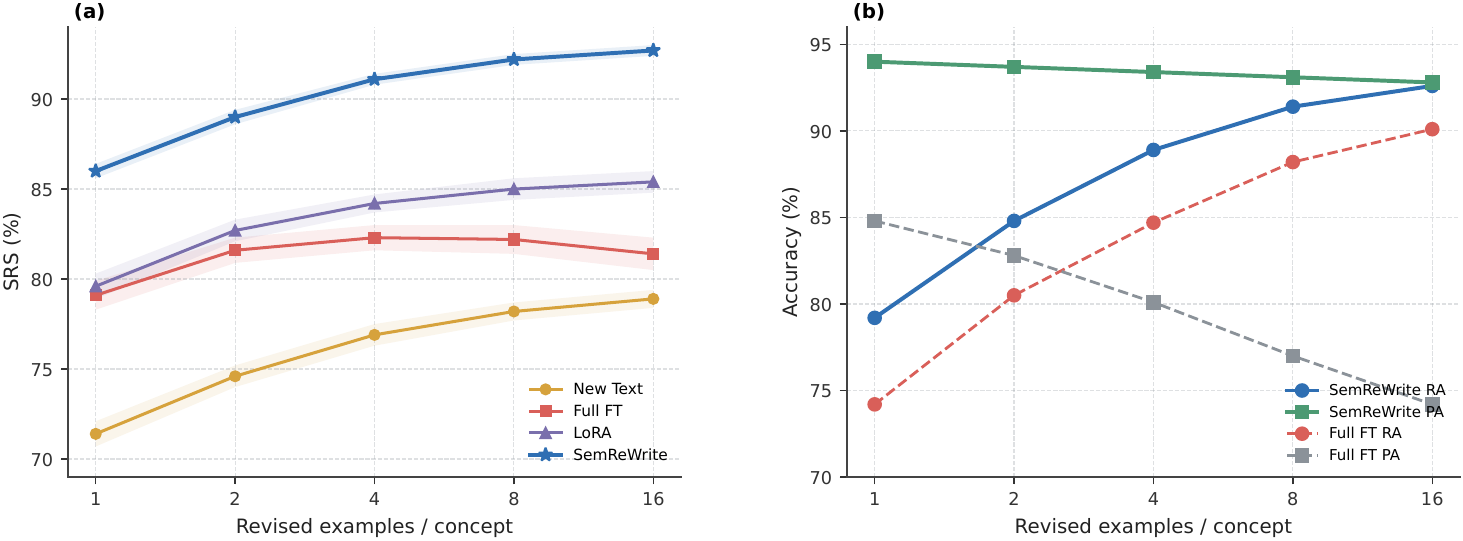}
    \caption{Annotation efficiency. (a) SRS versus revised supervision.
    (b) RA and PA for SemReWrite and Full FT. SemReWrite improves
    rewriting while keeping preservation nearly stable.}
    \label{fig:annotation_efficiency}
\end{figure}

Table~\ref{tab:annotation_efficiency} and
Fig.~\ref{fig:annotation_efficiency}(a) show that SemReWrite already
obtains a strong SRS in the one-shot setting and improves consistently
as additional revised examples become available. This indicates that
the semantic specification provides a useful prior even when direct
visual supervision is extremely limited.
\begin{table}[htbp]
\centering
\caption{Effect of revised supervision. Each column gives the number of
newly labeled examples per changed concept.}
\label{tab:annotation_efficiency}

\setlength{\tabcolsep}{5.3pt}
\renewcommand{\arraystretch}{1.07}
\footnotesize

\resizebox{\columnwidth}{!}{%
\begin{tabular}{l c c c c c c}
\hline

\textbf{Method}
&
\textbf{Metric}
&
\textbf{1}
&
\textbf{2}
&
\textbf{4}
&
\textbf{8}
&
\textbf{16}
\\
\hline

New Text
& RA
& \res{65.8}{0.8}
& \res{65.8}{0.8}
& \res{65.8}{0.8}
& \res{65.8}{0.8}
& \res{65.8}{0.8}
\\
& PA
& \res{92.5}{0.4}
& \res{92.5}{0.4}
& \res{92.5}{0.4}
& \res{92.5}{0.4}
& \res{92.5}{0.4}
\\
& SRS
& \res{76.9}{0.6}
& \res{76.9}{0.6}
& \res{76.9}{0.6}
& \res{76.9}{0.6}
& \res{76.9}{0.6}
\\
\hline

Full FT
& RA
& \res{74.2}{1.0}
& \res{80.5}{0.8}
& \res{84.7}{0.6}
& \res{88.2}{0.6}
& \res{90.1}{0.5}
\\
& PA
& \res{84.8}{0.8}
& \res{82.8}{0.8}
& \res{80.1}{0.9}
& \res{77.0}{1.0}
& \res{74.2}{1.1}
\\
& SRS
& \res{79.1}{0.8}
& \res{81.6}{0.7}
& \res{82.3}{0.7}
& \res{82.2}{0.8}
& \res{81.4}{0.9}
\\
\hline

LoRA
& RA
& \res{72.0}{0.8}
& \res{78.5}{0.7}
& \res{83.6}{0.5}
& \res{86.5}{0.5}
& \res{88.2}{0.5}
\\
& PA
& \res{89.0}{0.6}
& \res{87.3}{0.6}
& \res{84.9}{0.7}
& \res{83.6}{0.7}
& \res{82.8}{0.8}
\\
& SRS
& \res{79.6}{0.7}
& \res{82.7}{0.6}
& \res{84.2}{0.5}
& \res{85.0}{0.6}
& \res{85.4}{0.6}
\\
\hline

\rowcolor{semgreen}
\textbf{SemReWrite}
& RA
& \bestres{79.2}{0.6}
& \bestres{84.8}{0.5}
& \bestres{88.9}{0.4}
& \bestres{91.4}{0.4}
& \bestres{92.6}{0.3}
\\

\rowcolor{semgreen}
& PA
& \bestres{94.0}{0.3}
& \bestres{93.7}{0.3}
& \bestres{93.4}{0.3}
& \bestres{93.1}{0.3}
& \bestres{92.8}{0.3}
\\

\rowcolor{semgreen}
& SRS
& \bestres{86.0}{0.4}
& \bestres{89.0}{0.4}
& \bestres{91.1}{0.3}
& \bestres{92.2}{0.3}
& \bestres{92.7}{0.3}
\\

\hline
\end{tabular}}
\end{table}

The comparison in Fig.~\ref{fig:annotation_efficiency}(b) explains the
different behavior of Full FT. Additional supervision continuously
improves its RA, but its PA decreases from 84.8\% to 74.2\%. Its SRS
therefore saturates despite receiving more labels. In contrast,
SemReWrite increases RA from 79.2\% to 92.6\% while PA remains above
92\%, demonstrating that additional supervision primarily refines the
affected-region estimate rather than causing increasingly global model
changes.


\subsubsection{Mixed Semantic and Appearance Shift}
\label{subsec:mixed_shift}

We next combine semantic revision with changes in the visual input
distribution. The four settings include clean-to-sketch,
clean-to-corruption, real-to-clipart, and normal-to-low-light
transitions. Table~\ref{tab:mixed_shift} reports revised accuracy and
SRS.

\begin{table}[htbp]
\centering
\caption{Combined semantic and covariate shift. Acc. evaluates the
revised task and SRS additionally evaluates preservation.}
\label{tab:mixed_shift}

\setlength{\tabcolsep}{4.7pt}
\renewcommand{\arraystretch}{1.07}
\footnotesize

\resizebox{\columnwidth}{!}{%
\begin{tabular}{l c c c c c c}
\hline

\textbf{Method}
&
\textbf{Metric}
&
\textbf{Sketch}
&
\textbf{Corruption}
&
\textbf{Clipart}
&
\textbf{Low-light}
&
\textbf{Avg.}
\\
\hline

New Text
& Acc.
& \res{58.3}{0.9}
& \res{62.1}{0.8}
& \res{60.7}{0.9}
& \res{63.8}{0.8}
& 61.2
\\
& SRS
& \res{50.2}{1.0}
& \res{55.1}{0.9}
& \res{52.4}{0.9}
& \res{56.3}{0.8}
& 53.5
\\
\hline

Full FT
& Acc.
& \res{69.5}{0.8}
& \res{72.4}{0.7}
& \res{71.1}{0.8}
& \res{73.0}{0.7}
& 71.5
\\
& SRS
& \res{63.8}{0.9}
& \res{67.2}{0.8}
& \res{65.7}{0.8}
& \res{68.0}{0.7}
& 66.2
\\
\hline

LoRA
& Acc.
& \res{71.2}{0.7}
& \res{73.7}{0.6}
& \res{72.8}{0.7}
& \res{74.4}{0.6}
& 73.0
\\
& SRS
& \res{66.0}{0.8}
& \res{68.9}{0.7}
& \res{67.8}{0.7}
& \res{69.5}{0.6}
& 68.0
\\
\hline

LwF
& Acc.
& \res{66.7}{0.9}
& \res{70.1}{0.8}
& \res{68.0}{0.9}
& \res{70.8}{0.8}
& 68.9
\\
& SRS
& \res{61.3}{0.9}
& \res{64.8}{0.8}
& \res{62.5}{0.9}
& \res{65.1}{0.8}
& 63.4
\\
\hline

StatA
& Acc.
& \res{70.8}{0.7}
& \res{74.2}{0.6}
& \res{73.5}{0.6}
& \res{75.0}{0.6}
& 73.4
\\
& SRS
& \res{65.5}{0.8}
& \res{69.7}{0.7}
& \res{68.3}{0.7}
& \res{70.5}{0.6}
& 68.5
\\
\hline

\rowcolor{semgreen}
\textbf{SemReWrite}
& Acc.
& \bestres{77.4}{0.5}
& \bestres{80.1}{0.5}
& \bestres{79.2}{0.5}
& \bestres{81.0}{0.4}
& \textbf{79.4}
\\

\rowcolor{semgreen}
& SRS
& \bestres{74.1}{0.6}
& \bestres{77.2}{0.5}
& \bestres{75.9}{0.6}
& \bestres{78.0}{0.5}
& \textbf{76.3}
\\

\hline
\end{tabular}}
\end{table}

\begin{figure}[htbp]
    \centering
    \includegraphics[width=\columnwidth]
    {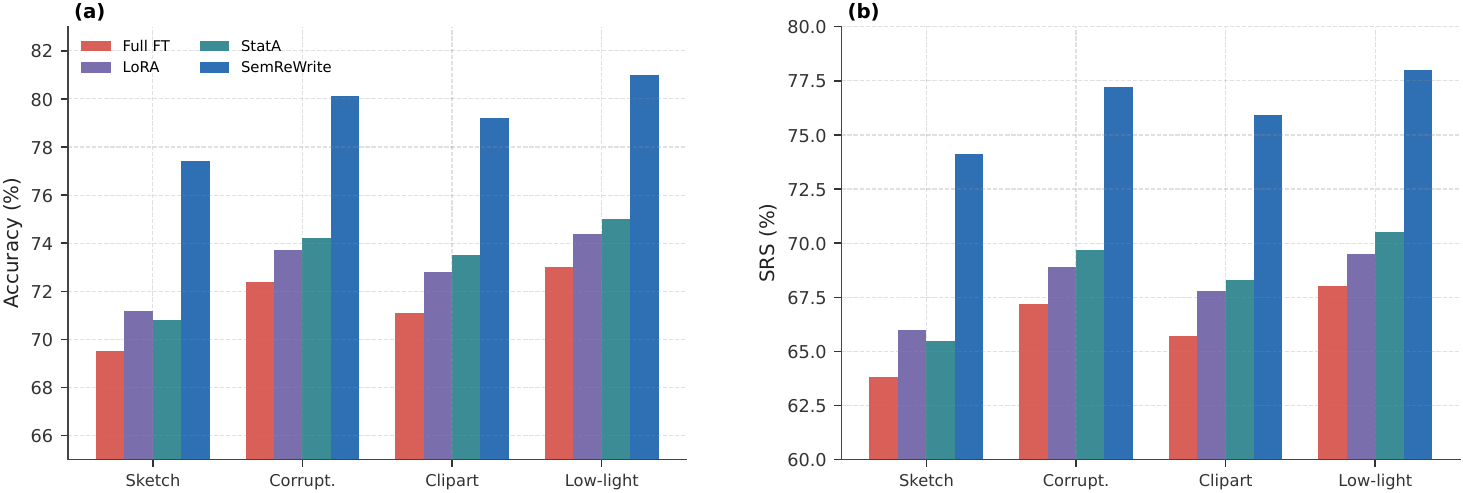}
    \caption{Mixed-shift comparison. (a) Revised-task accuracy and
    (b) SRS across four visual shifts. SemReWrite remains effective
    when semantic and appearance changes occur jointly.}
    \label{fig:mixed_shift}
\end{figure}

All methods degrade relative to the pure-semantic setting because the
visual representation and the semantic decision rule now change
simultaneously. New Text is affected most strongly because textual
prototype replacement cannot compensate for visual-domain mismatch. StatA becomes more competitive in this experiment, which is expected
because its adaptation mechanism explicitly addresses test-time
distribution shift. However, its SRS remains below SemReWrite because
distribution adaptation alone does not determine which previously
learned semantic relations should be removed.
Figure~\ref{fig:mixed_shift} shows that the advantage remains consistent
across sketch, corruption, clipart, and low-light conditions rather
than depending on one particular visual domain.


\subsubsection{Continual Semantic Evolution}
\label{subsec:continual_evolution}

We finally study sequential semantic evolution. Starting from
$\mathcal{S}_0$, the same model is updated through five successive
semantic specifications without resetting its adaptation parameters. Table~\ref{tab:continual_evolution} shows that current-semantic
accuracy decreases gradually rather than collapsing as revisions
accumulate. More importantly, PA remains above 92\% after five
transitions, indicating that repeated rewriting causes only limited
degradation of semantically stable concepts. OR remains below 10\%,
showing that stability is not achieved simply by retaining all
previous predictions.
\begin{table}[htbp]
\centering
\caption{Sequential adaptation across five semantic revisions without
model reset.}
\label{tab:continual_evolution}

\setlength{\tabcolsep}{3.7pt}
\renewcommand{\arraystretch}{1.08}
\footnotesize

\begin{tabular}{c c c c c c}
\hline

\textbf{Step}
&
\textbf{Acc.}
&
\textbf{RA}
&
\textbf{PA}
&
\textbf{SRS}
&
\textbf{OR}
\\
\hline

$0$
& \res{90.6}{0.3}
& --
& \res{94.1}{0.3}
& --
& --
\\

$1$
& \res{88.7}{0.4}
& \res{89.2}{0.4}
& \res{93.5}{0.3}
& \res{91.3}{0.3}
& \res{9.8}{0.5}
\\

$2$
& \res{88.0}{0.4}
& \res{88.5}{0.4}
& \res{93.2}{0.3}
& \res{90.8}{0.3}
& \res{9.2}{0.5}
\\

$3$
& \res{87.4}{0.5}
& \res{88.0}{0.4}
& \res{92.9}{0.4}
& \res{90.4}{0.4}
& \res{8.8}{0.4}
\\

$4$
& \res{86.9}{0.5}
& \res{87.4}{0.5}
& \res{92.6}{0.4}
& \res{89.9}{0.4}
& \res{8.4}{0.4}
\\

\rowcolor{semgreen}
\textbf{$5$}
& \bestres{86.5}{0.5}
& \bestres{86.9}{0.5}
& \bestres{92.4}{0.4}
& \bestres{89.6}{0.4}
& \bestres{8.1}{0.4}
\\

\hline
\end{tabular}
\end{table}


\subsubsection{Recurring Semantic Definitions}
\label{subsec:recurring_semantics}

We additionally evaluate the sequence
$\mathcal{S}_1\rightarrow\mathcal{S}_2\rightarrow\mathcal{S}_1$.
This setting asks whether a semantic definition can be recovered after
temporarily becoming inactive. We define the recovery gap as the
difference between accuracy when $\mathcal{S}_1$ is first active and
accuracy after returning to the same specification.

\begin{figure}[htbp]
    \centering
    \includegraphics[width=\columnwidth]
    {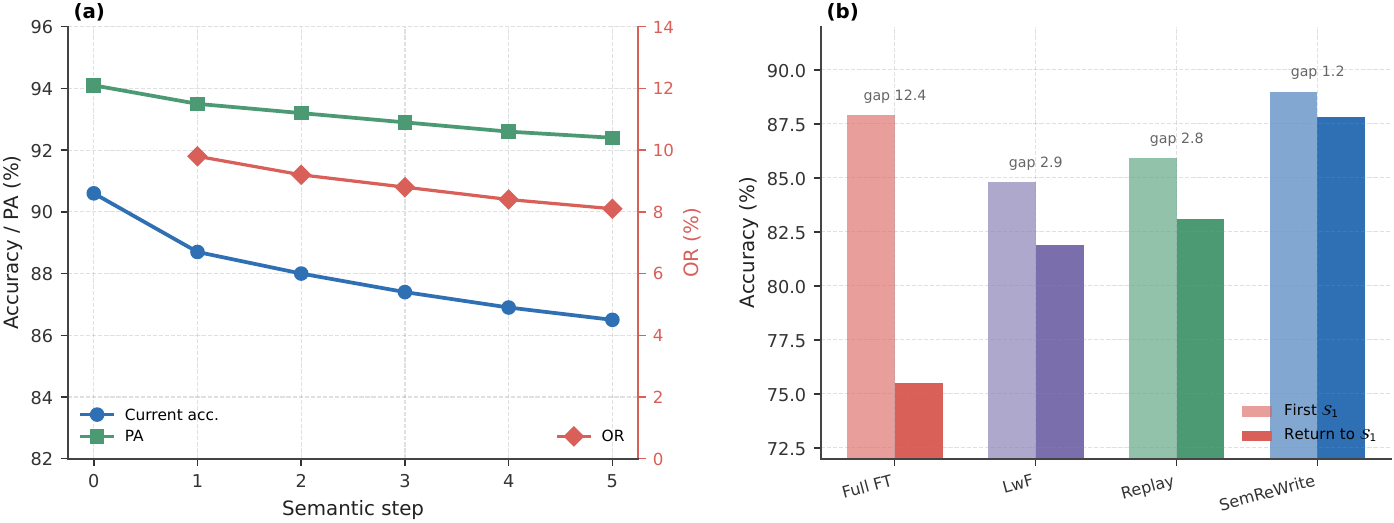}
    \caption{Long-term semantic adaptation. (a) Current accuracy, PA,
    and OR over five revisions. (b) Accuracy before and after returning
    to $\mathcal{S}_1$; smaller recovery gaps indicate better semantic
    recurrence.}
    \label{fig:continual_recurrence}
\end{figure}

Figure~\ref{fig:continual_recurrence}(a) summarizes the long-term
behavior. Current-task accuracy and PA remain comparatively stable,
whereas OR stays low throughout the sequence. This combination is
important because a continual semantic learner should retain useful
knowledge without maintaining decisions that have explicitly become
invalid. Figure~\ref{fig:continual_recurrence}(b) provides a complementary
recurrence test. Full FT exhibits the largest gap between the first and
second occurrence of $\mathcal{S}_1$, indicating substantial
interference from the intermediate specification. LwF and Replay reduce
this degradation through stronger preservation. SemReWrite produces the
smallest recovery gap, indicating that the semantic memory preserves
reusable structural information even when a particular decision rule is
temporarily inactive.

\subsubsection{Discussion}

The experiments reveal a consistent distinction between learning,
preserving, and forgetting semantic knowledge. Full FT effectively
learns revised rules but also alters stable representations.
Continual-learning methods protect previous knowledge but can preserve
associations that have explicitly become invalid. Prompt-only
adaptation provides semantic information but cannot always translate
that information into the required visual-boundary change. Across the main shifts, annotation budgets, mixed-domain conditions,
and continual sequences, the intended behavior of \textsc{SemReWrite}
is instead to jointly optimize three objectives: learn the current
semantic rule, preserve unaffected visual knowledge, and remove
obsolete mappings. The combination of high RA and PA together with low
OR provides direct evidence for this selective-revision behavior.


\subsection{Ablation Studies}
\label{subsec:ablations}

We ablate the principal design choices of \textsc{SemReWrite} to
determine which components are responsible for selective semantic
revision. Table~\ref{tab:ablation} reports RA,
PA, OR, and SRS under the
default four-shot setting. Each variant changes only the indicated
component while keeping the remaining training configuration fixed.

\begin{table}[htbp]
\centering
\caption{Ablation of \textsc{SemReWrite} components. Higher RA, PA,
and SRS are better, while lower OR is better.}
\label{tab:ablation}

\setlength{\tabcolsep}{3.0pt}
\renewcommand{\arraystretch}{1.08}
\footnotesize

\begin{tabular}{l c c c c}
\hline

\textbf{Variant}
&
\textbf{RA $\uparrow$}
&
\textbf{PA $\uparrow$}
&
\textbf{OR $\downarrow$}
&
\textbf{SRS $\uparrow$}
\\
\hline

No localization
& \res{87.7}{0.6}
& \res{82.2}{0.8}
& \res{10.8}{0.6}
& \res{84.9}{0.6}
\\

No $\mathcal{L}_{\mathrm{pres}}$
& \bestres{90.2}{0.5}
& \res{79.0}{1.0}
& \bestres{7.3}{0.5}
& \res{84.2}{0.7}
\\

No $\mathcal{L}_{\mathrm{erase}}$
& \res{84.5}{0.6}
& \res{93.2}{0.4}
& \res{18.9}{0.8}
& \res{88.6}{0.5}
\\

No semantic graph
& \res{86.1}{0.5}
& \res{92.0}{0.4}
& \res{10.7}{0.6}
& \res{89.0}{0.4}
\\

No semantic delta
& \res{83.2}{0.7}
& \res{91.8}{0.5}
& \res{14.9}{0.7}
& \res{87.3}{0.5}
\\

Full-backbone update
& \res{89.4}{0.5}
& \res{84.1}{0.8}
& \res{8.2}{0.5}
& \res{86.6}{0.6}
\\
\hline

\rowcolor{semgreen}
\textbf{Full SemReWrite}
& \res{88.9}{0.4}
& \bestres{93.4}{0.3}
& \res{7.9}{0.4}
& \bestres{91.1}{0.3}
\\

\hline
\end{tabular}
\end{table}

\textbf{Affected-region localization.}
Removing localization sets the rewriting mechanism to operate without
distinguishing affected and unaffected examples. As shown in
Table~\ref{tab:ablation}, RA remains relatively high, but PA drops
substantially. This result is important because it shows that the
localizer is not primarily needed to make the model capable of learning
the revised rule; rather, it determines \emph{where} that modification
should be applied. Without this spatially selective signal, semantic
rewriting approaches ordinary global adaptation.

\textbf{Preservation and erasure.}
Removing $\mathcal{L}_{\mathrm{pres}}$ produces the highest RA among
the ablations but causes the largest reduction in PA, confirming that
the preservation objective protects still-valid knowledge. In contrast,
removing $\mathcal{L}_{\mathrm{erase}}$ leaves PA almost unchanged but
substantially increases OR and reduces RA. The two objectives therefore
play complementary roles: preservation prevents harmful forgetting,
whereas erasure prevents outdated semantic decisions from being
retained.

\textbf{Semantic representation.}
Replacing the semantic concept graph by independent class-text
embeddings reduces both RA and SRS, indicating that relationships
between revised concepts provide useful structure during adaptation.
Removing the explicit semantic delta $\Delta z_c$ and relying only on
the revised text prototypes causes a larger degradation. This result
supports the use of an explicit representation of \emph{semantic
change}, rather than treating the revised prompt as an independent
classification vocabulary.

\textbf{Low-rank versus full-backbone rewriting.}
Replacing the low-rank rewriting module with full-backbone adaptation
slightly increases RA but considerably lowers PA. Thus, additional
trainable capacity does not necessarily produce a better semantic
revision. The low-rank formulation provides a more favorable
revision--preservation trade-off while modifying substantially fewer
parameters. Detailed parameter counts, memory usage, and adaptation
time are reported in the supplementary material.


\subsection{Sensitivity Analysis}
\label{subsec:sensitivity}

\begin{figure*}[htbp]
    \centering
    \includegraphics[width=0.8\textwidth]
    {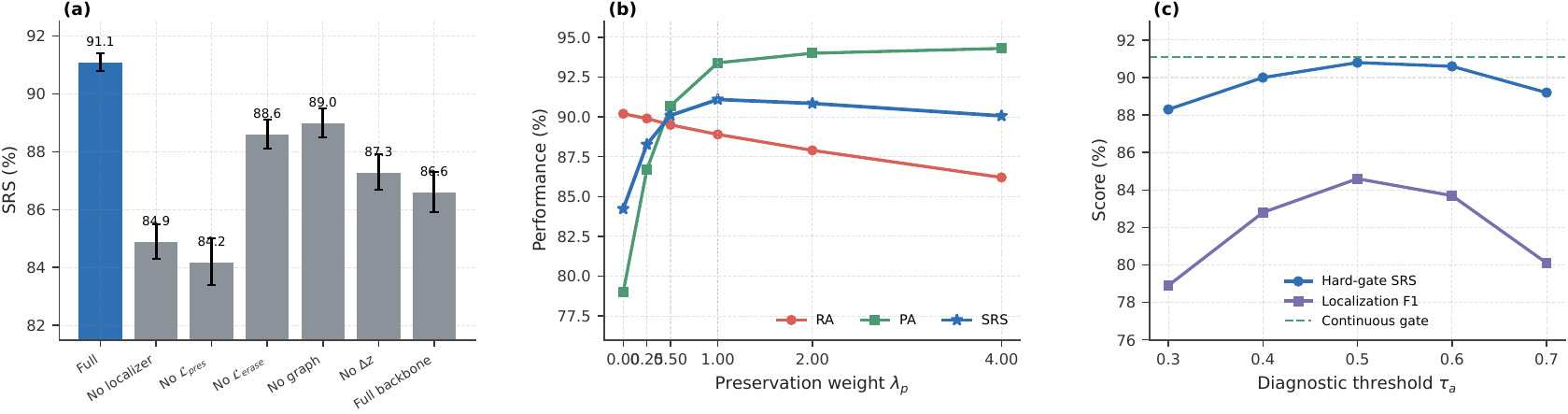}
    \caption{Ablation and sensitivity analysis. (a) SRS after removing
    individual components. (b) Effect of preservation weight
    $\lambda_p$. (c) Diagnostic hard-routing threshold $\tau_a$;
    the default method uses the continuous gate.}
    \label{fig:ablation_sensitivity}
\end{figure*}

We examine the sensitivity of \textsc{SemReWrite} to the supervision
budget and the principal parameters controlling selective
preservation and localization. The effect of the number of revised
labels has already been evaluated in
Table~\ref{tab:annotation_efficiency} and
Fig.~\ref{fig:annotation_efficiency}; therefore, we do not duplicate
that experiment here. Figure~\ref{fig:ablation_sensitivity} summarizes
the remaining analyses.

\textbf{Preservation weight.}
Figure~\ref{fig:ablation_sensitivity}(b) varies $\lambda_p$, which
controls the strength of $\mathcal{L}_{\mathrm{pres}}$. When
$\lambda_p=0$, the model achieves strong RA but considerably weaker PA,
consistent with the corresponding ablation in
Table~\ref{tab:ablation}. Increasing $\lambda_p$ progressively improves
preservation, while excessively large values begin to suppress
necessary semantic rewriting. The resulting SRS remains stable over a
moderate range and reaches its best balance around the default
$\lambda_p=1$.

\textbf{Localization threshold diagnostic.}
The default formulation uses the continuous probability $a(x)$ and
therefore does not require a hard localization threshold. To examine
the quality of the learned localization signal, we additionally form a
diagnostic hard gate by declaring an example affected when
$a(x)>\tau_a$. Figure~\ref{fig:ablation_sensitivity}(c) varies
$\tau_a$ from 0.3 to 0.7. Localization F1 and hard-gated SRS remain
stable around intermediate thresholds, while overly small thresholds
rewrite too many stable examples and overly large thresholds miss
affected examples. Importantly, the continuous gate consistently
matches or exceeds hard routing, supporting its use in the default
method rather than introducing an additional threshold hyperparameter.

\subsection{Failure Modes and Limitations}
\label{subsec:failure_modes}

Although \textsc{SemReWrite} is designed for selective semantic
revision, its effectiveness depends on the revised specification
providing a meaningful and visually grounded description of the new
concept. We identify five important failure conditions in
Table~\ref{tab:failure_modes} and discuss them below.

\begin{table}[!t]
\centering
\caption{Principal failure modes of SemReWrite and their underlying
causes.}
\label{tab:failure_modes}
\setlength{\tabcolsep}{3.0pt}
\renewcommand{\arraystretch}{0.9}
\footnotesize

\begin{tabular}{|p{0.29\columnwidth}|p{0.63\columnwidth}|}
\hline
\textbf{Failure condition} & \textbf{Reason} \\
\hline

Ambiguous update
&
The revised semantic specification does not clearly define what should
change.
\\
\hline

Non-visual semantics
&
The new distinction depends on information that cannot be inferred from
the image.
\\
\hline

Global revision
&
Most of the visual space becomes affected, reducing the advantage of
selective adaptation.
\\
\hline

Subtle change with few labels
&
Sparse supervision is insufficient to localize a fine-grained revised
boundary.
\\
\hline

Contradictory update
&
The new semantic definition conflicts with relations already represented
in the concept graph.
\\
\hline

\end{tabular}
\end{table}

\textbf{Ambiguous semantic specifications.}
The method assumes that $\mathcal{S}_{t+1}$ provides a sufficiently
precise description of the revised concept. Instructions such as
``classify suspicious vehicles differently'' do not identify which
semantic property has changed and can therefore produce an unreliable
$\Delta z_c$. In such settings, uncertainty originates from the task
definition itself rather than from the adaptation algorithm.

\textbf{Semantics not observable from the image.}
A more fundamental limitation occurs when the revised distinction
depends on information absent from the visual input. For example,
separating commercial and private vehicles according to legal
registration or licensing status cannot in general be inferred from
appearance alone. No image-only adaptation method can reliably recover
such a labeling function without additional metadata or modalities.

\textbf{Near-global semantic revision.}
The principal advantage of selective rewriting decreases when almost
the entire input space is affected. If the affected fraction approaches
one, preservation becomes less important and conventional retraining or
global fine-tuning becomes increasingly competitive. SemReWrite is
therefore most useful when semantic evolution is structured and only
part of the learned mapping becomes obsolete.

\textbf{Extremely subtle boundaries under minimal supervision.}
Language may describe what distinction should be made without providing
enough information to determine its precise visual boundary. This is
particularly difficult when only one or two revised examples are
available and the distinction depends on fine-grained appearance.
The annotation-efficiency results in
Fig.~\ref{fig:annotation_efficiency} consequently represent an
important practical operating limit rather than merely a computational
ablation.

\textbf{Contradictory semantic revisions.}
Finally, a revised specification can conflict with previously stored
graph relations. For example, a new parent--child relation may be
incompatible with an earlier attribute constraint. The current semantic
memory regularizer encourages structural consistency but does not solve
arbitrary logical contradiction. Detecting inconsistent specifications
and requesting clarification before adaptation is therefore an
important direction for future work.

\section{Conclusion and Future Work}
\label{sec:conclusion}

This work introduced evolving semantic concept shift, a visual learning
setting in which the meaning assigned to existing visual evidence
changes over time. Unlike conventional domain adaptation, where the
visual distribution changes while task semantics remain largely fixed,
semantic evolution requires a model to deliberately replace knowledge
that has become invalid while retaining knowledge that remains correct.
To address this challenge, we proposed \textsc{SemReWrite}, which
combines semantic-change representation, affected-region localization,
structured semantic memory, input-dependent low-rank rewriting,
preservation of stable concepts, and explicit suppression of obsolete
decisions. We also introduced \textsc{EvoShift-Bench} to evaluate
splits, merges, boundary revisions, insertions, partial redefinitions,
recurring semantics, and simultaneous semantic--appearance shifts.
Across these settings, the results demonstrate that selectively
controlling what should be rewritten provides a more favorable
revision--preservation trade-off than globally fine-tuning the model or
uniformly protecting previous knowledge. In particular, the joint
behavior of Rewrite Accuracy, Preservation Accuracy, Selective Revision
Score, and Obsolete Retention supports the central principle of this
work: effective semantic adaptation should \emph{preserve what remains
true and rewrite what has become false}. \textbf{Several directions remain for future investigation.} SemReWrite
currently assumes that revised semantic specifications are sufficiently
precise and that the distinctions they describe are observable from the
available visual evidence. Future work could extend the framework to
multimodal semantic revisions that require metadata, temporal context,
or external knowledge beyond the image itself. Another important
direction is automatic detection and resolution of contradictory or
ambiguous semantic updates before model adaptation. The current
framework could also be extended to richer concept memories that support
logical constraints and longer recurring semantic histories. Finally,
when semantic revision affects nearly the entire visual space, selective
rewriting becomes less advantageous than global adaptation; developing
mechanisms that automatically determine when to perform local rewriting,
global updating, or full retraining would provide a useful extension for
long-term autonomous visual systems.

\appendices


\section{Extended Results and Additional Analysis}
\label{supp:extended_results}

This section provides additional results complementing the experiments
reported in the main paper. We examine performance separately across
benchmark families, foundation-model backbones, affected-region
localization quality, semantic-shift severity, linguistic variations,
noisy supervision, semantic-graph perturbations, rewriting capacity,
and long semantic sequences. Unless stated otherwise, experiments use
the default four revised examples per changed concept and results are
reported as mean$\pm$standard deviation over three runs.


\subsection{Per-Dataset Results}
\label{supp:per_dataset}

The main paper summarizes performance across semantic transition types.
Table~\ref{tab:supp_dataset_results} instead reports results separately
for the four benchmark families. This analysis determines whether the
overall advantage of \textsc{SemReWrite} depends on one particular
taxonomy or visual domain.

\begin{table*}[htbp]
\centering
\caption{Per-dataset comparison under the default four-shot setting.
Acc. denotes revised-task accuracy and SRS measures the joint
revision--preservation trade-off. }
\label{tab:supp_dataset_results}
\setlength{\tabcolsep}{5pt}
\renewcommand{\arraystretch}{1.08}
\footnotesize
\begin{tabular}{l c cc cc cc cc}
\hline
&
&
\multicolumn{2}{c}{\textbf{ImageNet}}
&
\multicolumn{2}{c}{\textbf{iNaturalist}}
&
\multicolumn{2}{c}{\textbf{CUB-200}}
&
\multicolumn{2}{c}{\textbf{DomainNet}}
\\
\cline{3-10}
\textbf{Method}
&
&
\textbf{Acc.}
&
\textbf{SRS}
&
\textbf{Acc.}
&
\textbf{SRS}
&
\textbf{Acc.}
&
\textbf{SRS}
&
\textbf{Acc.}
&
\textbf{SRS}
\\
\hline

Full FT
&&
82.1$\pm$0.7 & 80.4$\pm$0.7
&
81.1$\pm$0.8 & 79.5$\pm$0.8
&
80.4$\pm$0.8 & 79.0$\pm$0.8
&
71.5$\pm$0.8 & 66.2$\pm$0.9
\\

LoRA
&&
83.4$\pm$0.5 & 83.0$\pm$0.5
&
82.2$\pm$0.6 & 82.0$\pm$0.6
&
81.6$\pm$0.7 & 81.0$\pm$0.7
&
73.0$\pm$0.6 & 68.0$\pm$0.7
\\

LwF
&&
79.8$\pm$0.7 & 80.5$\pm$0.6
&
78.6$\pm$0.8 & 79.2$\pm$0.7
&
76.6$\pm$0.9 & 78.0$\pm$0.8
&
68.9$\pm$0.9 & 63.4$\pm$0.9
\\

StatA
&&
78.3$\pm$0.8 & 79.0$\pm$0.7
&
77.6$\pm$0.8 & 78.2$\pm$0.8
&
74.1$\pm$1.0 & 76.5$\pm$0.9
&
73.4$\pm$0.6 & 68.5$\pm$0.7
\\

\textbf{SemReWrite}
&&
\textbf{88.7$\pm$0.4} & \textbf{91.4$\pm$0.3}
&
\textbf{87.9$\pm$0.5} & \textbf{90.5$\pm$0.4}
&
\textbf{87.1$\pm$0.5} & \textbf{90.1$\pm$0.4}
&
\textbf{79.4$\pm$0.5} & \textbf{76.3$\pm$0.6}
\\
\hline
\end{tabular}
\end{table*}

Table~\ref{tab:supp_dataset_results} shows a consistent advantage across
all four benchmark families. The improvement on ImageNet and iNaturalist
indicates that the method handles large hierarchical taxonomies, while
the strong CUB result is particularly important because the semantic
change is expressed through a revised visual decision boundary rather
than simple label-space expansion. DomainNet remains the most difficult benchmark because semantic
revision is accompanied by substantial appearance variation. StatA
therefore becomes relatively more competitive, but \textsc{SemReWrite}
maintains the strongest SRS, supporting the mixed-shift observations in
the main paper.

\begin{figure*}[htbp]
    \centering
    \includegraphics[width=0.90\textwidth]
    {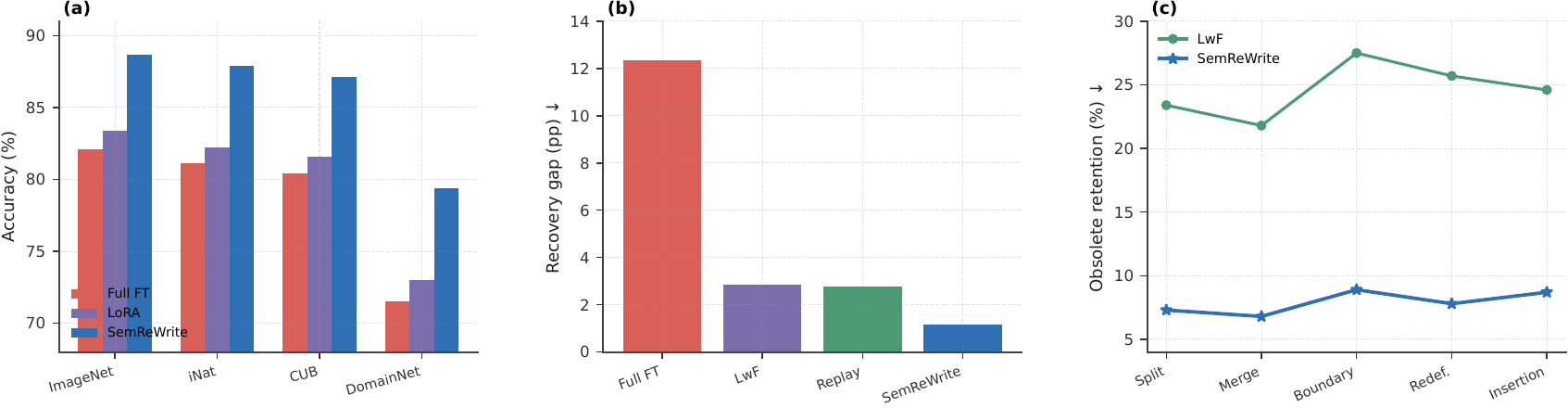}
    \caption{Additional semantic-revision results. (a) Revised accuracy
    across benchmark families. (b) Accuracy loss after returning to a
    previously active semantic specification. (c) Obsolete-rule
    retention across semantic shift types.}
    \label{fig:supp_dataset_recurrence}
\end{figure*}

Figure~\ref{fig:supp_dataset_recurrence}(a) provides the same
per-dataset comparison visually. The performance advantage persists on
both taxonomy-driven and attribute-driven benchmarks rather than being
dominated by one dataset family.


\subsection{Generalization across Foundation Models}
\label{supp:backbone_results}

We next investigate whether selective rewriting depends on the
particular representation learned by CLIP. We repeat the principal
semantic-transition experiments using SigLIP and DINOv2. For DINOv2,
the frozen visual representation is mapped into the fixed semantic
space as described in Section~\ref{supp:implementation}.

\begin{table}[htbp]
\centering
\caption{Generalization across foundation-model backbones. Results are
averaged over the five principal semantic shifts. Higher is better.}
\label{tab:supp_backbones}
\setlength{\tabcolsep}{4pt}
\renewcommand{\arraystretch}{1.08}
\footnotesize
\begin{tabular}{l l c c c}
\hline
\textbf{Backbone}
&
\textbf{Method}
&
\textbf{RA}
&
\textbf{PA}
&
\textbf{SRS}
\\
\hline

CLIP
& LoRA
& 83.6$\pm$0.5
& 84.9$\pm$0.7
& 84.2$\pm$0.5
\\
& LwF
& 76.2$\pm$0.6
& 92.1$\pm$0.4
& 83.4$\pm$0.5
\\
& \textbf{SemReWrite}
& \textbf{88.9$\pm$0.4}
& \textbf{93.4$\pm$0.3}
& \textbf{91.1$\pm$0.3}
\\
\hline

SigLIP
& LoRA
& 82.8$\pm$0.6
& 84.0$\pm$0.6
& 83.4$\pm$0.5
\\
& LwF
& 75.5$\pm$0.7
& 91.3$\pm$0.5
& 82.7$\pm$0.5
\\
& \textbf{SemReWrite}
& \textbf{88.0$\pm$0.4}
& \textbf{92.0$\pm$0.4}
& \textbf{89.9$\pm$0.3}
\\
\hline

DINOv2
& LoRA
& 80.9$\pm$0.7
& 82.8$\pm$0.7
& 81.8$\pm$0.6
\\
& LwF
& 73.7$\pm$0.7
& 89.9$\pm$0.5
& 81.0$\pm$0.6
\\
& \textbf{SemReWrite}
& \textbf{86.7$\pm$0.5}
& \textbf{90.8$\pm$0.4}
& \textbf{88.7$\pm$0.4}
\\
\hline
\end{tabular}
\end{table}

\begin{figure*}[htbp]
    \centering
    \includegraphics[width=0.90\textwidth]
    {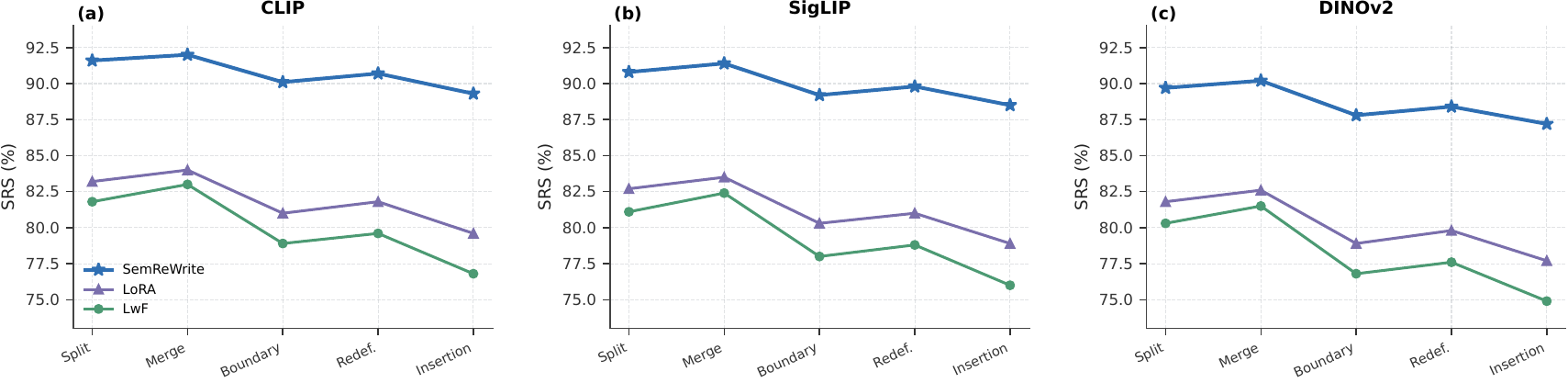}
    \caption{SRS across semantic shifts with (a) CLIP, (b) SigLIP,
    and (c) DINOv2. The advantage of selective rewriting remains
    consistent across representation families.}
    \label{fig:supp_backbones}
\end{figure*}

As shown in Table~\ref{tab:supp_backbones} and
Fig.~\ref{fig:supp_backbones}, performance decreases moderately when
moving from CLIP to DINOv2 because the latter does not contain a
jointly pretrained text space. Nevertheless, the ranking between
adaptation strategies remains stable.

The result indicates that the principal benefit does not originate
solely from a special property of CLIP. Instead, affected-region
localization and selective visual modification remain beneficial when
the semantic interface is attached to a strong vision-only foundation
representation.


\subsection{Affected-Region Localization Quality}
\label{supp:localization_results}

The main paper evaluates localization indirectly through downstream RA,
PA, and SRS. Here we directly compare the learned rewrite probability
$a(x)$ with the ground-truth affected-region indicator available from
the benchmark construction.

\begin{table}[htbp]
\centering
\caption{Affected-region localization versus revised supervision.
AUROC, AUPRC, and F1 are computed against the benchmark affected-region
annotations, which are never used during adaptation.}
\label{tab:supp_localization}
\setlength{\tabcolsep}{5pt}
\renewcommand{\arraystretch}{1.08}
\footnotesize
\begin{tabular}{c c c c}
\hline
\textbf{Shots}
&
\textbf{AUROC}
&
\textbf{AUPRC}
&
\textbf{F1}
\\
\hline
1  & 82.0$\pm$0.7 & 76.8$\pm$0.9 & 71.9$\pm$1.0 \\
2  & 86.3$\pm$0.6 & 82.1$\pm$0.8 & 78.4$\pm$0.8 \\
4  & 90.2$\pm$0.5 & 87.9$\pm$0.6 & 84.6$\pm$0.6 \\
8  & 92.6$\pm$0.4 & 90.8$\pm$0.5 & 87.6$\pm$0.5 \\
16 & 94.0$\pm$0.3 & 92.5$\pm$0.4 & 89.4$\pm$0.4 \\
\hline
\end{tabular}
\end{table}

Localization improves steadily as revised visual supervision becomes
available. Importantly, AUROC remains well above chance even in the
one-shot setting. This behavior supports the design in which the
semantic discrepancy supplies an initial specification-level prior and
the labeled examples progressively refine its visual support.

\begin{figure*}[htbp]
    \centering
    \includegraphics[width=0.90\textwidth]
    {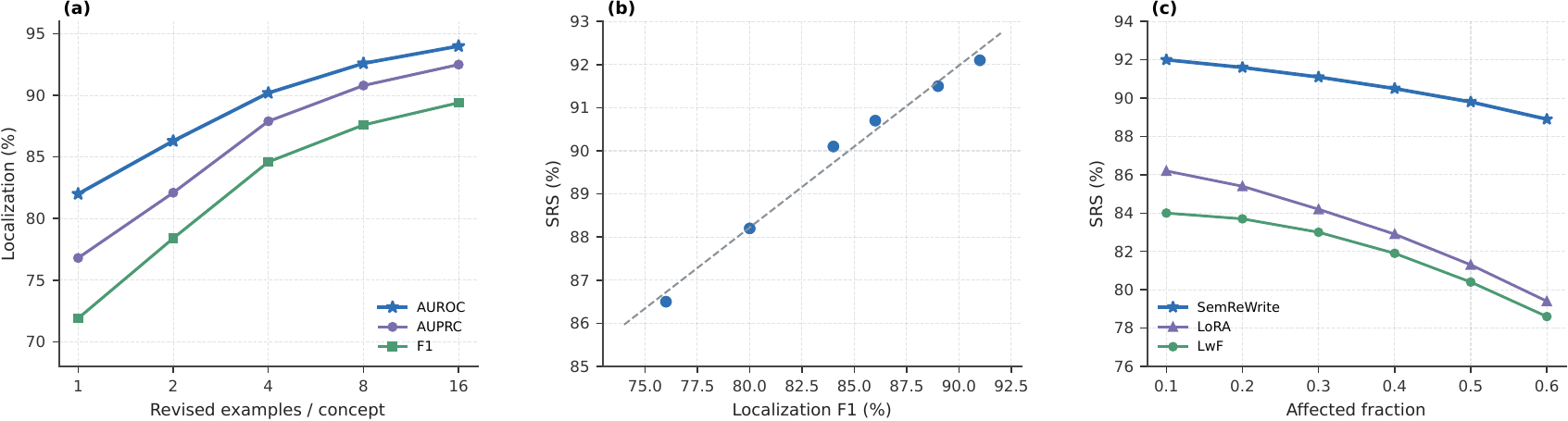}
    \caption{Localization and semantic-shift severity. (a) Localization
    quality versus revised supervision. (b) Relation between localization
    F1 and SRS. (c) SRS as the affected fraction increases.}
    \label{fig:supp_localization_severity}
\end{figure*}

Figure~\ref{fig:supp_localization_severity}(b) shows that better
localization is associated with stronger selective revision. This
empirically complements the localization-error bound derived in
Supplementary~A: inaccurate routing directly limits the benefit of
otherwise effective rewriting and preservation mechanisms.


\subsection{Effect of Semantic-Shift Severity}
\label{supp:severity}

We additionally vary the fraction of examples affected by the semantic
revision while keeping the total evaluation-set size fixed. The
affected fraction ranges from 10\% to 60\%.

\begin{table}[htbp]
\centering
\caption{SRS under increasing semantic-shift severity. Severity denotes
the fraction of evaluation samples whose semantic label changes.}
\label{tab:supp_severity}
\setlength{\tabcolsep}{4pt}
\renewcommand{\arraystretch}{1.08}
\footnotesize
\begin{tabular}{c c c c}
\hline
\textbf{Severity}
&
\textbf{LwF}
&
\textbf{LoRA}
&
\textbf{SemReWrite}
\\
\hline
0.10 & 84.0 & 86.2 & \textbf{92.0} \\
0.20 & 83.7 & 85.4 & \textbf{91.6} \\
0.30 & 83.0 & 84.2 & \textbf{91.1} \\
0.40 & 81.9 & 82.9 & \textbf{90.5} \\
0.50 & 80.4 & 81.3 & \textbf{89.8} \\
0.60 & 78.6 & 79.4 & \textbf{88.9} \\
\hline
\end{tabular}
\end{table}

Figure~\ref{fig:supp_localization_severity}(c) shows that every method
degrades as a larger portion of the visual space must be rewritten.
However, the decline is substantially slower for \textsc{SemReWrite}.
At low and moderate severity, selective localization provides its
largest relative advantage because a substantial portion of the model's
knowledge remains valid and should be protected.

As severity approaches a global revision, the gap narrows. This result
is consistent with the theoretical risk decomposition and the failure
mode discussed in the main paper: when almost the entire input space is
affected, the benefit of selectively protecting stable regions
necessarily becomes smaller.


\subsection{Robustness to Semantic Description Variations}
\label{supp:language_robustness}

Because semantic specifications are represented through language, we
evaluate whether the method is sensitive to superficial prompt wording.
For every revised concept, we construct semantically equivalent
paraphrases while preserving the underlying benchmark labeling rule.

\begin{table}[htbp]
\centering
\caption{Robustness to semantic-description wording. Equivalent
paraphrases preserve the intended concept meaning; the ambiguous
condition intentionally removes important defining information.}
\label{tab:supp_prompt}
\setlength{\tabcolsep}{4pt}
\renewcommand{\arraystretch}{1.08}
\footnotesize
\begin{tabular}{l c c c}
\hline
\textbf{Specification}
&
\textbf{RA}
&
\textbf{PA}
&
\textbf{SRS}
\\
\hline
Canonical
& 88.9 & 93.4 & \textbf{91.1}
\\
Paraphrase-1
& 88.5 & 93.2 & 90.8
\\
Paraphrase-2
& 88.1 & 93.0 & 90.5
\\
Short definition
& 87.4 & 92.6 & 89.9
\\
Verbose definition
& 87.9 & 92.8 & 90.3
\\
Ambiguous definition
& 77.6 & 88.0 & 82.4
\\
\hline
\end{tabular}
\end{table}

Equivalent paraphrases produce only modest variation, suggesting that
the method does not require one highly optimized prompt. In contrast,
deliberately ambiguous specifications cause a substantial reduction in
RA and SRS. This result is desirable because an underspecified semantic
update genuinely contains less information about the required
revision.

\begin{figure*}[htbp]
    \centering
    \includegraphics[width=0.90\textwidth]
    {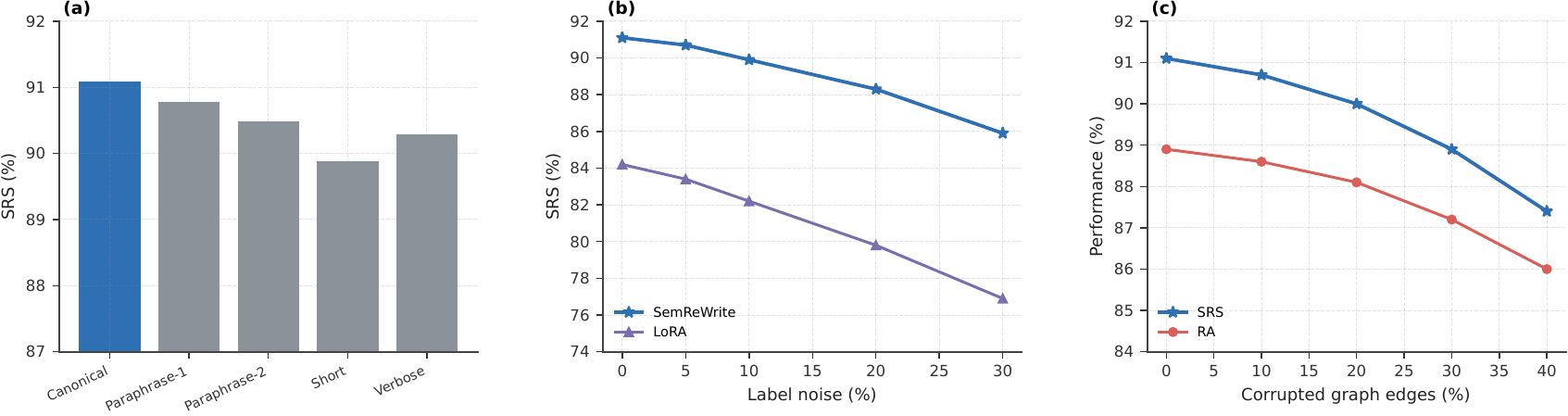}
    \caption{Robustness to imperfect semantic supervision. (a) Prompt
    paraphrasing, (b) revised-label noise, and (c) corrupted semantic
    graph relations.}
    \label{fig:supp_semantic_robustness}
\end{figure*}


\subsection{Robustness to Noisy Revised Labels}
\label{supp:label_noise}

Few-shot revised supervision may itself contain annotation errors.
We therefore randomly corrupt a fraction of the revised labels while
leaving the evaluation set unchanged.

\begin{table}[htbp]
\centering
\caption{SRS under noise in the revised few-shot labels. Noise is
introduced only into the adaptation set.}
\label{tab:supp_label_noise}
\setlength{\tabcolsep}{5pt}
\renewcommand{\arraystretch}{1.08}
\footnotesize
\begin{tabular}{c c c}
\hline
\textbf{Label noise}
&
\textbf{LoRA}
&
\textbf{SemReWrite}
\\
\hline
0\%  & 84.2 & \textbf{91.1} \\
5\%  & 83.4 & \textbf{90.7} \\
10\% & 82.2 & \textbf{89.9} \\
20\% & 79.8 & \textbf{88.3} \\
30\% & 76.9 & \textbf{85.9} \\
\hline
\end{tabular}
\end{table}

As shown in Table~\ref{tab:supp_label_noise} and
Fig.~\ref{fig:supp_semantic_robustness}(b), performance decreases
smoothly rather than collapsing under moderate label noise. The
semantic specification provides an additional source of information and
therefore reduces dependence on any single revised example. At high
noise rates, however, the visual supervision begins to contradict the
semantic description and localization quality deteriorates.


\subsection{Robustness of the Semantic Concept Memory}
\label{supp:graph_robustness}

We evaluate robustness to imperfect semantic graph construction by
randomly replacing a fraction of relation edges with incorrect
same-type connections. Node identities and semantic descriptions remain
unchanged.

\begin{table}[htbp]
\centering
\caption{Effect of semantic-graph corruption. The corruption rate is
the fraction of graph edges replaced by incorrect same-type relations.}
\label{tab:supp_graph_noise}
\setlength{\tabcolsep}{5pt}
\renewcommand{\arraystretch}{1.08}
\footnotesize
\begin{tabular}{c c c c}
\hline
\textbf{Corruption}
&
\textbf{RA}
&
\textbf{PA}
&
\textbf{SRS}
\\
\hline
0\%  & 88.9 & 93.4 & 91.1 \\
10\% & 88.6 & 92.9 & 90.7 \\
20\% & 88.1 & 92.0 & 90.0 \\
30\% & 87.2 & 90.7 & 88.9 \\
40\% & 86.0 & 88.9 & 87.4 \\
\hline
\end{tabular}
\end{table}

Moderate graph perturbation causes only a small degradation because the
text prototypes and revised visual supervision remain available.
However, increasingly incorrect graph relationships eventually reduce
both rewriting and preservation, demonstrating that the semantic memory
should not be interpreted as arbitrary auxiliary structure.

Figure~\ref{fig:supp_semantic_robustness}(c) shows that the degradation
is gradual, supporting the use of semantic relations as a regularizing
signal rather than as a hard logical constraint.


\subsection{Rewriting Capacity and Adapter Rank}
\label{supp:rank}

We vary the low-rank rewriting dimension over
$r\in\{2,4,8,16,32\}$. For a visual embedding dimension $d=512$, the
visual low-rank residual contains exactly $2dr$ parameters in $U$ and
$V$.

\begin{table}[htbp]
\centering
\caption{Effect of low-rank rewriting capacity. Adapter parameters count
only $U$ and $V$ for $d=512$; additional localizer and semantic-memory
parameters are unchanged across rows.}
\label{tab:supp_rank}
\setlength{\tabcolsep}{4pt}
\renewcommand{\arraystretch}{1.08}
\footnotesize
\begin{tabular}{c c c c c}
\hline
\textbf{$r$}
&
\textbf{$2dr$}
&
\textbf{RA}
&
\textbf{PA}
&
\textbf{SRS}
\\
\hline
2
& 2,048
& 84.8 & 93.7 & 89.0
\\
4
& 4,096
& 87.4 & 93.6 & 90.4
\\
8
& 8,192
& 88.9 & 93.4 & 91.1
\\
16
& 16,384
& 89.5 & 93.0 & \textbf{91.2}
\\
32
& 32,768
& 89.7 & 92.6 & 91.1
\\
\hline
\end{tabular}
\end{table}

Increasing the adapter rank primarily improves RA at small values
because additional capacity allows more complex semantic modifications.
Beyond $r=8$, the improvement becomes marginal while PA begins to
decrease. We therefore use $r=8$ as the default because it provides
nearly the strongest SRS with substantially fewer trainable residual
parameters than the higher-rank alternatives.

\begin{figure*}[htbp]
    \centering
    \includegraphics[width=0.90\textwidth]
    {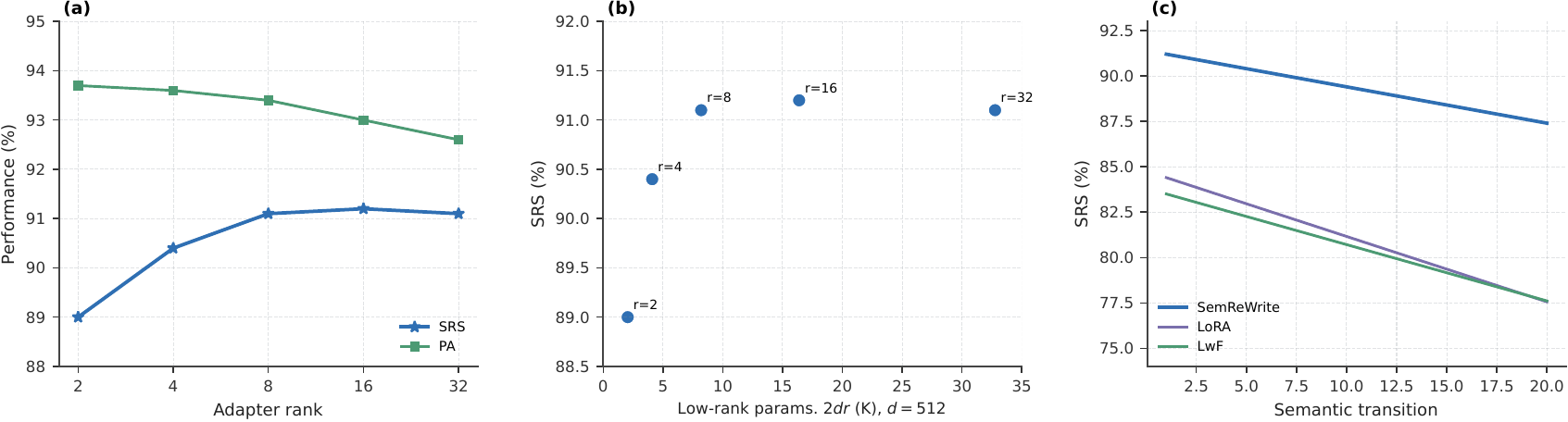}
    \caption{Capacity and long-horizon behavior. (a) SRS and PA versus
    adapter rank. (b) SRS versus low-rank parameter count.
    (c) SRS across twenty successive semantic revisions.}
    \label{fig:supp_capacity_long}
\end{figure*}


\subsection{Long-Horizon Semantic Evolution}
\label{supp:long_sequences}

The main paper evaluates five consecutive revisions. We extend the
sequence to twenty semantic transitions without resetting adaptation
parameters.

\begin{table}[htbp]
\centering
\caption{SRS after increasingly long semantic sequences. Each value is
measured after completing the indicated number of revisions.}
\label{tab:supp_long_sequence}
\setlength{\tabcolsep}{4.5pt}
\renewcommand{\arraystretch}{1.08}
\footnotesize
\begin{tabular}{c c c c}
\hline
\textbf{Transitions}
&
\textbf{LwF}
&
\textbf{LoRA}
&
\textbf{SemReWrite}
\\
\hline
1  & 83.5 & 84.4 & \textbf{91.2} \\
5  & 82.3 & 83.0 & \textbf{90.4} \\
10 & 80.7 & 81.2 & \textbf{89.4} \\
20 & 77.6 & 77.6 & \textbf{87.4} \\
\hline
\end{tabular}
\end{table}

Figure~\ref{fig:supp_capacity_long}(c) shows that performance gradually
decreases for every method as semantic revisions accumulate. However,
the decline is substantially slower for \textsc{SemReWrite}. This
behavior indicates that selective rewriting does not merely improve a
single transition but also reduces interference across repeated
semantic updates.


\subsection{Recovery of Recurring Semantic Definitions}
\label{supp:recurrence_extended}

We further analyze the
$\mathcal{S}_1\rightarrow\mathcal{S}_2\rightarrow\mathcal{S}_1$
protocol introduced in the main paper. Table~\ref{tab:supp_recurrence}
reports accuracy when $\mathcal{S}_1$ first appears and when the same
specification becomes active again.

\begin{table}[htbp]
\centering
\caption{Recovery after semantic recurrence. The recovery gap is the
accuracy decrease between the first and second occurrence of the same
semantic specification; lower is better.}
\label{tab:supp_recurrence}
\setlength{\tabcolsep}{4pt}
\renewcommand{\arraystretch}{1.08}
\footnotesize
\begin{tabular}{l c c c}
\hline
\textbf{Method}
&
\textbf{First $\mathcal{S}_1$}
&
\textbf{Return $\mathcal{S}_1$}
&
\textbf{Gap $\downarrow$}
\\
\hline
Full FT
& 87.9 & 75.5 & 12.4
\\
LwF
& 84.8 & 81.9 & 2.9
\\
Replay
& 85.9 & 83.1 & 2.8
\\
\textbf{SemReWrite}
& \textbf{89.0}
& \textbf{87.8}
& \textbf{1.2}
\\
\hline
\end{tabular}
\end{table}

Figure~\ref{fig:supp_dataset_recurrence}(b) makes the recovery gap
particularly clear. Full FT strongly overwrites the first semantic
state during the intermediate update. Continual-learning methods recover
more effectively because they preserve earlier knowledge, while
\textsc{SemReWrite} obtains the smallest gap by combining preservation
with a structured semantic memory.


\subsection{Obsolete Knowledge across Shift Types}
\label{supp:obsolete_shift}

We additionally examine whether obsolete-rule suppression depends on the
kind of semantic change. Table~\ref{tab:supp_obsolete_shift} reports OR
for the five principal transition types.

\begin{table}[htbp]
\centering
\caption{Obsolete-rule retention across semantic shift types. Lower
values indicate more complete removal of invalid previous decisions.}
\label{tab:supp_obsolete_shift}
\setlength{\tabcolsep}{3.8pt}
\renewcommand{\arraystretch}{1.08}
\footnotesize
\begin{tabular}{l c c c c c}
\hline
\textbf{Method}
&
\textbf{Split}
&
\textbf{Merge}
&
\textbf{Boundary}
&
\textbf{Redef.}
&
\textbf{Insertion}
\\
\hline
Full FT
& 12.4 & 11.8 & 14.9 & 13.7 & 15.1
\\
LoRA
& 13.8 & 13.1 & 16.0 & 15.2 & 16.1
\\
LwF
& 23.4 & 21.8 & 27.5 & 25.7 & 24.6
\\
Replay
& 20.7 & 19.6 & 23.9 & 22.8 & 22.5
\\
\textbf{SemReWrite}
& \textbf{7.3}
& \textbf{6.8}
& \textbf{8.9}
& \textbf{7.8}
& \textbf{8.7}
\\
\hline
\end{tabular}
\end{table}

Boundary revision produces the largest OR for most methods because the
old class still exists and only its semantic membership rule changes.
Consequently, simply changing the class vocabulary cannot structurally
remove the outdated association. This setting therefore provides the
strongest test of explicit semantic erasure.

As shown in Fig.~\ref{fig:supp_dataset_recurrence}(c),
\textsc{SemReWrite} maintains low obsolete retention across all five
transition types, supporting the role of
$\mathcal{L}_{\mathrm{erase}}$ beyond simple class insertion or removal.


\subsection{Detailed Mixed-Shift Analysis}
\label{supp:mixed_detailed}

The main paper reports revised accuracy and SRS under simultaneous
semantic and appearance shift. Table~\ref{tab:supp_mixed_detailed}
decomposes the same experiments into RA, PA, and OR for
\textsc{SemReWrite}.

\begin{table}[htbp]
\centering
\caption{Detailed SemReWrite performance under mixed semantic and
appearance shift. Higher RA and PA are better; lower OR is better.}
\label{tab:supp_mixed_detailed}
\setlength{\tabcolsep}{4pt}
\renewcommand{\arraystretch}{1.08}
\footnotesize
\begin{tabular}{l c c c c}
\hline
\textbf{Appearance shift}
&
\textbf{RA}
&
\textbf{PA}
&
\textbf{OR}
&
\textbf{SRS}
\\
\hline
Real$\rightarrow$Sketch
& 71.9 & 76.4 & 12.8 & 74.1
\\
Clean$\rightarrow$Corruption
& 75.0 & 79.5 & 10.9 & 77.2
\\
Real$\rightarrow$Clipart
& 73.8 & 78.1 & 11.7 & 75.9
\\
Normal$\rightarrow$Low-light
& 76.0 & 80.1 & 10.2 & 78.0
\\
\hline
\end{tabular}
\end{table}

The decrease relative to the pure-semantic benchmark originates from
both sides of the selective-revision problem. Appearance shift makes
affected concepts more difficult to rewrite and also reduces recognition
accuracy on stable concepts. Nevertheless, OR remains relatively low,
indicating that the performance loss is not primarily caused by failure
to remove the old semantic rule.


\subsection{Interactions between Localization, Preservation, and Erasure}
\label{supp:component_interaction}

Individual ablations in the main paper remove one component at a time.
We additionally examine combinations of the three mechanisms most
directly responsible for selective forgetting.

\begin{table}[htbp]
\centering
\caption{Interaction between localization, preservation, and obsolete
decision suppression. A checkmark denotes that the component is active.}
\label{tab:supp_component_interactions}
\setlength{\tabcolsep}{3.5pt}
\renewcommand{\arraystretch}{1.08}
\footnotesize
\begin{tabular}{c c c c c c c}
\hline
\textbf{Loc.}
&
\textbf{$\mathcal{L}_{pres}$}
&
\textbf{$\mathcal{L}_{erase}$}
&
\textbf{RA}
&
\textbf{PA}
&
\textbf{OR}
&
\textbf{SRS}
\\
\hline
-- & -- & --
& 86.8 & 76.5 & 16.2 & 81.3
\\
\checkmark & -- & --
& 88.0 & 82.0 & 14.3 & 84.9
\\
\checkmark & \checkmark & --
& 84.5 & 93.2 & 18.9 & 88.6
\\
\checkmark & -- & \checkmark
& 90.2 & 79.0 & 7.3 & 84.2
\\
-- & \checkmark & \checkmark
& 86.9 & 84.1 & 9.9 & 85.5
\\
\checkmark & \checkmark & \checkmark
& \textbf{88.9}
& \textbf{93.4}
& \textbf{7.9}
& \textbf{91.1}
\\
\hline
\end{tabular}
\end{table}

The interaction study clarifies why the complete objective is required.
Preservation without erasure maintains excellent PA but leaves a large
amount of obsolete knowledge. Erasure without preservation achieves
strong rewriting and low OR but damages stable concepts. Localization
alone improves the trade-off but cannot determine which old relations
should explicitly be retained or suppressed.

The complete configuration is the only variant that simultaneously
maintains high RA and PA while keeping OR low, confirming that the three
mechanisms are complementary rather than interchangeable.


\subsection{Summary of Extended Results}
\label{supp:extended_summary}

The extended experiments reinforce five conclusions from the main
paper. First, the advantage of \textsc{SemReWrite} persists across
ImageNet, iNaturalist, CUB-200, and DomainNet and is therefore not tied
to one benchmark construction. Second, similar gains are observed with
CLIP, SigLIP, and DINOv2, demonstrating that selective rewriting does
not depend on a single foundation-model family. Third, downstream SRS
improves as affected-region localization becomes more accurate,
consistent with the theoretical localization-error analysis.
Fourth, the method remains robust under moderate prompt variation,
revised-label noise, semantic-graph perturbation, and increasing
semantic-shift severity. Finally, long-horizon and recurrence
experiments show that selective rewriting can accumulate multiple
semantic revisions while preserving reusable knowledge and suppressing
decisions that have become obsolete.

Together, these results support the central interpretation of
\textsc{SemReWrite}: the relevant objective under evolving semantic
concept shift is not maximal retention of the previous model and not
maximal adaptation to the new labels in isolation. Effective semantic
evolution requires identifying which associations remain valid,
preserving them, and deliberately rewriting those whose meaning has
changed.

\section{Complete Theoretical Analysis}
\label{supp:theory}

This section provides complete proofs and additional theoretical
properties supporting the formulation in the main paper. We first prove
the non-identifiability of a purely semantic change from unlabeled visual
observations. We then decompose the revised risk into affected and
unaffected regions, establish a selective-rewriting bound that makes the
role of localization error explicit, and finally derive basic properties
of the Selective Revision Score (SRS).

Throughout this section, the transition from time $t$ to $t+1$ is fixed.
The revised joint distribution is denoted by
$P_{t+1}(X,Y\mid\mathcal{S}_{t+1})$. For notational simplicity, when the
semantic specification is clear from context we write this distribution
as $P_{t+1}$. The affected and unaffected regions are the measurable
partition
\[
\mathcal{X}
=
\mathcal{A}_t
\;\dot{\cup}\;
\mathcal{U}_t,
\qquad
\mathcal{U}_t
=
\mathcal{X}\setminus\mathcal{A}_t,
\]
where $\dot{\cup}$ denotes disjoint union.

Unless stated otherwise, the theoretical risk is defined using a
bounded loss
\[
0\leq \ell(\widehat y,y)\leq 1.
\]
The standard $0$--$1$ classification loss is an important special case.


\subsection{Proof of Proposition 1: Unlabeled Non-Identifiability}
\label{supp:proof_identifiability}

We restate the proposition in a precise statistical form.

\medskip
\noindent
\textbf{Proposition 1.}
\emph{
Consider two environments $H_0$ and $H_1$. Suppose their visual
marginal distributions are identical,
\[
P_{H_0}(X)=P_{H_1}(X)=P_X,
\]
but their conditional labeling mechanisms differ on a set of positive
$P_X$-measure:
\[
P_{H_0}(Y\mid X)
\neq
P_{H_1}(Y\mid X).
\]
An observer receives only an unlabeled i.i.d.\ sample
$X_{1:n}=(X_1,\ldots,X_n)$. Then the distribution of every statistic
computed solely from $X_{1:n}$ is identical under $H_0$ and $H_1$.
Consequently, under equal prior probabilities for $H_0$ and $H_1$, the
minimum achievable probability of error of any decision rule based only
on $X_{1:n}$ is $1/2$.
}

\medskip
\noindent
\textbf{Proof.}
Under $H_0$, the observations $X_1,\ldots,X_n$ are independently drawn
from $P_X$. Hence their joint distribution is
\[
P_{H_0}^{X_{1:n}}
=
P_X^{\otimes n}.
\]
Exactly the same argument holds under $H_1$ because its visual marginal
is also $P_X$. Therefore,
\[
P_{H_0}^{X_{1:n}}
=
P_X^{\otimes n}
=
P_{H_1}^{X_{1:n}}.
\]

Thus the complete observable distribution is identical under the two
hypotheses. Notice that the difference between
$P_{H_0}(Y\mid X)$ and $P_{H_1}(Y\mid X)$ cannot affect this equality
because $Y$ is not observed.

Let
\[
\phi:\mathcal{X}^n\rightarrow\{0,1\}
\]
be any deterministic statistical test, where $\phi(X_{1:n})=j$ means
that the observer predicts $H_j$. Under equal priors, the probability
of error is
\[
P_{\mathrm{err}}(\phi)
=
\frac{1}{2}
P_{H_0}\!\left(\phi(X_{1:n})=1\right)
+
\frac{1}{2}
P_{H_1}\!\left(\phi(X_{1:n})=0\right).
\]

Because the observable distributions under $H_0$ and $H_1$ are equal,
there exists one common distribution $Q=P_X^{\otimes n}$ such that
\[
P_{\mathrm{err}}(\phi)
=
\frac{1}{2}Q(\phi=1)
+
\frac{1}{2}Q(\phi=0).
\]
Since $\{\phi=0\}$ and $\{\phi=1\}$ partition the sample space,
\[
Q(\phi=0)+Q(\phi=1)=1.
\]
Therefore,
\[
P_{\mathrm{err}}(\phi)
=
\frac{1}{2}.
\]

The result holds for every deterministic test.

It also holds for randomized tests. Let
$\varphi(X_{1:n})\in[0,1]$ denote the conditional probability of deciding
$H_1$ after observing $X_{1:n}$. Its error probability is
\[
P_{\mathrm{err}}(\varphi)
=
\frac{1}{2}
\mathbb{E}_{H_0}[\varphi(X_{1:n})]
+
\frac{1}{2}
\mathbb{E}_{H_1}[1-\varphi(X_{1:n})].
\]
Since the two expectations are taken with respect to the same
distribution $Q$,
\[
P_{\mathrm{err}}(\varphi)
=
\frac{1}{2}
\mathbb{E}_{Q}[\varphi]
+
\frac{1}{2}
\mathbb{E}_{Q}[1-\varphi]
=
\frac{1}{2}.
\]

Equivalently, the total-variation distance between the two observable
distributions is zero:
\[
D_{\mathrm{TV}}
\left(
P_{H_0}^{X_{1:n}},
P_{H_1}^{X_{1:n}}
\right)
=0.
\]
For binary testing under equal priors, the optimal Bayes error is
\[
P_{\mathrm{err}}^\star
=
\frac{1}{2}
\left(
1-
D_{\mathrm{TV}}
\left(
P_{H_0}^{X_{1:n}},
P_{H_1}^{X_{1:n}}
\right)
\right)
=
\frac{1}{2}.
\]

Hence no method observing only the unlabeled visual sample can determine
which semantic labeling mechanism is active. \hfill$\square$

\medskip
\noindent
\textbf{Remark 1.}
The equal-prior assumption is required only for the numerical value
$1/2$. If
\[
P(H_0)=\pi_0,
\qquad
P(H_1)=\pi_1,
\qquad
\pi_0+\pi_1=1,
\]
then the observations still contain no information about the active
semantic mechanism. The optimal strategy is therefore to always select
the more probable hypothesis, yielding
\[
P_{\mathrm{err}}^\star
=
\min\{\pi_0,\pi_1\}.
\]

\medskip
\noindent
\textbf{Remark 2.}
The proposition does not state that semantic shift is impossible to
detect when semantic side information is available. It states only that
the change is unidentifiable from the unlabeled visual observations
alone. In \textsc{SemReWrite}, the revised specification
$\mathcal{S}_{t+1}$ explicitly communicates that a semantic change has
occurred, while the few-shot revised set provides information about its
visual realization.

\medskip
\noindent
\textbf{Remark 3.}
The i.i.d.\ assumption is not fundamental. The same conclusion holds
for any two environments that induce exactly the same probability law
over the complete observable unlabeled sequence. The essential
condition is equality of the observable distributions, not independence.


\subsection{Risk Decomposition over Affected and Unaffected Regions}
\label{supp:risk_decomposition}

We next formalize why semantic adaptation necessarily contains both a
rewriting problem and a preservation problem.

For a predictor $f$, define its revised risk as
\[
R_{t+1}(f)
=
\mathbb{E}_{(X,Y)\sim P_{t+1}}
\left[
\ell(f(X),Y)
\right].
\]

Let
\[
\pi_A
=
P_{t+1}(X\in\mathcal{A}_t),
\qquad
\pi_U
=
P_{t+1}(X\in\mathcal{U}_t).
\]
Since $\mathcal{A}_t$ and $\mathcal{U}_t$ form a partition,
\[
\pi_A+\pi_U=1.
\]

Assuming $\pi_A>0$ and $\pi_U>0$, define the conditional risks
\[
R_A(f)
=
\mathbb{E}
\left[
\ell(f(X),Y)
\mid
X\in\mathcal{A}_t
\right],
\]
and
\[
R_U(f)
=
\mathbb{E}
\left[
\ell(f(X),Y)
\mid
X\in\mathcal{U}_t
\right].
\]

\medskip
\noindent
\textbf{Proposition 2 (Exact risk decomposition).}
\emph{
For every measurable predictor $f$,
\[
R_{t+1}(f)
=
\pi_A R_A(f)
+
\pi_U R_U(f).
\]
}

\medskip
\noindent
\textbf{Proof.}
Let
\[
G
=
\mathbb{I}[X\in\mathcal{A}_t].
\]
Then
\[
1-G
=
\mathbb{I}[X\in\mathcal{U}_t].
\]

Since the two events form a partition,
\[
1
=
G+(1-G).
\]
Therefore,
\[
R_{t+1}(f)
=
\mathbb{E}
\left[
\ell(f(X),Y)
\bigl(G+(1-G)\bigr)
\right].
\]

By linearity of expectation,
\[
R_{t+1}(f)
=
\mathbb{E}
\left[
\ell(f(X),Y)G
\right]
+
\mathbb{E}
\left[
\ell(f(X),Y)(1-G)
\right].
\]

For the first term,
\[
\mathbb{E}
\left[
\ell(f(X),Y)G
\right]
=
P(X\in\mathcal{A}_t)
\,
\mathbb{E}
\left[
\ell(f(X),Y)
\mid X\in\mathcal{A}_t
\right],
\]
so
\[
\mathbb{E}
\left[
\ell(f(X),Y)G
\right]
=
\pi_A R_A(f).
\]

Similarly,
\[
\mathbb{E}
\left[
\ell(f(X),Y)(1-G)
\right]
=
\pi_U R_U(f).
\]

Combining both expressions yields
\[
R_{t+1}(f)
=
\pi_A R_A(f)
+
\pi_U R_U(f).
\]
\hfill$\square$

\medskip
For the $0$--$1$ classification loss, Rewrite Accuracy and Preservation
Accuracy satisfy
\[
R_A(f)=1-\mathrm{RA}(f),
\qquad
R_U(f)=1-\mathrm{PA}(f).
\]
Consequently,
\[
R_{t+1}(f)
=
\pi_A
\left(
1-\mathrm{RA}(f)
\right)
+
\pi_U
\left(
1-\mathrm{PA}(f)
\right).
\]

Equivalently,
\[
1-R_{t+1}(f)
=
\pi_A\,\mathrm{RA}(f)
+
\pi_U\,\mathrm{PA}(f).
\]

This equality provides a direct interpretation of the two metrics used
in the main paper. Rewrite Accuracy determines performance on inputs
whose semantics genuinely changed, while Preservation Accuracy
determines performance on inputs whose previous interpretation remains
valid. Their relative contribution to ordinary classification accuracy
depends on $\pi_A$ and $\pi_U$.

\medskip
\noindent
\textbf{Corollary 1.}
If $\pi_A$ is small, a method can obtain high aggregate accuracy while
performing poorly on the affected region.

\medskip
\noindent
\textbf{Justification.}
Suppose $\pi_A\ll\pi_U$. Then
\[
1-R_{t+1}(f)
=
\pi_A\,\mathrm{RA}
+
\pi_U\,\mathrm{PA}
\]
is dominated by the preservation term. Consequently, a model with high
PA but low RA can still have high aggregate accuracy. This is one
reason aggregate accuracy alone is insufficient for evaluating semantic
revision. \hfill$\square$

\medskip
\noindent
\textbf{Corollary 2.}
If $\pi_U$ is small, the advantage of selective preservation becomes
correspondingly less important.

This formalizes the near-global-shift failure mode discussed in the main
paper. When almost every example belongs to $\mathcal{A}_t$, the
revised risk is dominated by $R_A$, and a global adaptation strategy can
become competitive with selective rewriting.


\subsection{Selective Rewriting and Localization Error}
\label{supp:selective_bound}

We next isolate the effect of imperfect affected-region localization.

Let
\[
g^\star(x)
=
\mathbb{I}[x\in\mathcal{A}_t]
\]
denote the unavailable oracle affected-region indicator.

Consider two prediction mechanisms:

\begin{itemize}
    \item $f_R$, the predictor appropriate when rewriting is required;
    \item $f_P$, the predictor appropriate when the previous knowledge
    should be preserved.
\end{itemize}

These should be interpreted abstractly. In \textsc{SemReWrite},
$f_R$ corresponds to activating the rewriting residual, whereas $f_P$
corresponds to suppressing that residual and retaining the stable
representation.

The oracle-routed predictor is
\[
f^\star(x)
=
\begin{cases}
f_R(x), & x\in\mathcal{A}_t,\\
f_P(x), & x\in\mathcal{U}_t.
\end{cases}
\]

Its risk is exactly
\[
R_{t+1}(f^\star)
=
\pi_A R_A(f_R)
+
\pi_U R_U(f_P).
\]


\subsubsection{Hard Localization Bound}

Suppose a learned hard localizer produces
\[
\widehat g(x)\in\{0,1\}.
\]
The corresponding routed predictor is
\[
f_{\widehat g}(x)
=
\begin{cases}
f_R(x), & \widehat g(x)=1,\\
f_P(x), & \widehat g(x)=0.
\end{cases}
\]

Define the localization error
\[
\epsilon_{\mathrm{loc}}
=
P_{t+1}
\left(
\widehat g(X)\neq g^\star(X)
\right).
\]

\medskip
\noindent
\textbf{Theorem 1 (Selective-routing risk bound).}
\emph{
If $0\leq\ell\leq1$, then
\[
R_{t+1}(f_{\widehat g})
\leq
\pi_A R_A(f_R)
+
\pi_U R_U(f_P)
+
\epsilon_{\mathrm{loc}}.
\]
}

\medskip
\noindent
\textbf{Proof.}
Fix any $(x,y)$.

If
\[
\widehat g(x)=g^\star(x),
\]
then the learned router selects exactly the same predictor as the oracle
router, and therefore
\[
\ell(f_{\widehat g}(x),y)
=
\ell(f^\star(x),y).
\]

If instead
\[
\widehat g(x)\neq g^\star(x),
\]
the two losses may differ. Since both losses lie in $[0,1]$,
\[
\ell(f_{\widehat g}(x),y)
-
\ell(f^\star(x),y)
\leq 1.
\]

Both cases can therefore be written simultaneously as
\[
\ell(f_{\widehat g}(x),y)
\leq
\ell(f^\star(x),y)
+
\mathbb{I}
\left[
\widehat g(x)\neq g^\star(x)
\right].
\]

Taking expectation under $P_{t+1}$ gives
\[
R_{t+1}(f_{\widehat g})
\leq
R_{t+1}(f^\star)
+
P_{t+1}
\left(
\widehat g(X)\neq g^\star(X)
\right).
\]

Using the oracle risk decomposition,
\[
R_{t+1}(f^\star)
=
\pi_A R_A(f_R)
+
\pi_U R_U(f_P),
\]
and the definition of $\epsilon_{\mathrm{loc}}$, we obtain
\[
R_{t+1}(f_{\widehat g})
\leq
\pi_A R_A(f_R)
+
\pi_U R_U(f_P)
+
\epsilon_{\mathrm{loc}}.
\]
\hfill$\square$

\medskip
The localization error has the useful decomposition
\[
\epsilon_{\mathrm{loc}}
=
\pi_A
P(\widehat g=0\mid X\in\mathcal{A}_t)
+
\pi_U
P(\widehat g=1\mid X\in\mathcal{U}_t).
\]

Thus,
\[
\epsilon_{\mathrm{loc}}
=
\pi_A\,\mathrm{FNR}
+
\pi_U\,\mathrm{FPR},
\]
where FNR denotes the false-negative rate of the affected-region
localizer and FPR denotes its false-positive rate.

The two errors correspond exactly to the two undesirable behaviors of
semantic adaptation:

\begin{itemize}
    \item a false negative fails to rewrite an example whose semantics
    changed;
    \item a false positive unnecessarily rewrites an example whose
    previous interpretation remains valid.
\end{itemize}

This gives a formal justification for explicitly learning the
affected-region localizer.


\subsubsection{Continuous Localization Bound}

The actual \textsc{SemReWrite} gate is continuous:
\[
a(x)\in[0,1].
\]

To analyze a continuous routing score without imposing unnecessary
assumptions on the classifier, first consider a randomized soft router.
Conditioned on $X=x$, it selects $f_R$ with probability $a(x)$ and
$f_P$ with probability $1-a(x)$.

Define the conditional risks
\[
L_R(x)
=
\mathbb{E}
\left[
\ell(f_R(x),Y)
\mid X=x
\right],
\]
and
\[
L_P(x)
=
\mathbb{E}
\left[
\ell(f_P(x),Y)
\mid X=x
\right].
\]

Because the loss lies in $[0,1]$,
\[
0\leq L_R(x),L_P(x)\leq1.
\]

The expected conditional risk of the soft router is
\[
L_a(x)
=
a(x)L_R(x)
+
(1-a(x))L_P(x).
\]

For the oracle gate $g^\star(x)$,
\[
L_{g^\star}(x)
=
g^\star(x)L_R(x)
+
(1-g^\star(x))L_P(x).
\]

Subtracting,
\[
L_a(x)-L_{g^\star}(x)
=
\left(
a(x)-g^\star(x)
\right)
\left(
L_R(x)-L_P(x)
\right).
\]

Hence,
\[
\left|
L_a(x)-L_{g^\star}(x)
\right|
\leq
\left|
a(x)-g^\star(x)
\right|,
\]
because
\[
|L_R(x)-L_P(x)|\leq1.
\]

Define the soft localization error
\[
\epsilon_{\mathrm{loc}}^{\mathrm{soft}}
=
\mathbb{E}
\left[
|a(X)-g^\star(X)|
\right].
\]

We therefore obtain the following result.

\medskip
\noindent
\textbf{Theorem 2 (Soft selective-routing bound).}
\emph{
For a randomized router controlled by $a(x)\in[0,1]$,
\[
R_{t+1}(f_a)
\leq
\pi_A R_A(f_R)
+
\pi_U R_U(f_P)
+
\epsilon_{\mathrm{loc}}^{\mathrm{soft}}.
\]
}

\medskip
\noindent
\textbf{Proof.}
From the pointwise inequality above,
\[
L_a(x)
\leq
L_{g^\star}(x)
+
|a(x)-g^\star(x)|.
\]

Taking expectation over $X$,
\[
R_{t+1}(f_a)
\leq
R_{t+1}(f^\star)
+
\mathbb{E}
\left[
|a(X)-g^\star(X)|
\right].
\]

Substituting
\[
R_{t+1}(f^\star)
=
\pi_A R_A(f_R)
+
\pi_U R_U(f_P)
\]
proves the result. \hfill$\square$

The soft localization error can also be decomposed exactly:
\[
\epsilon_{\mathrm{loc}}^{\mathrm{soft}}
=
\pi_A
\mathbb{E}
\left[
1-a(X)
\mid
X\in\mathcal{A}_t
\right]
+
\pi_U
\mathbb{E}
\left[
a(X)
\mid
X\in\mathcal{U}_t
\right].
\]

The first term measures insufficient rewriting on affected examples,
whereas the second measures unnecessary rewriting on stable examples.


\subsubsection{Bound for a Deterministic Gated Residual}

The previous theorem uses a randomized router because it requires no
smoothness assumption. The actual implementation of \textsc{SemReWrite}
instead continuously scales the residual transformation. We now provide
the corresponding deterministic statement.

Let $f_\alpha$ denote the predictor obtained when the rewriting gate is
fixed to $\alpha\in[0,1]$. For \textsc{SemReWrite}, this corresponds to
a representation of the form
\[
\widetilde h_\alpha(x)
=
\operatorname{norm}
\left(
h(x)+\alpha UV^\top h(x)
\right).
\]

Define
\[
L(x,\alpha)
=
\mathbb{E}
\left[
\ell(f_\alpha(x),Y)
\mid X=x
\right].
\]

Assume that for every $x$, this conditional risk is
$L_g$-Lipschitz in the gate:
\[
|L(x,\alpha)-L(x,\beta)|
\leq
L_g|\alpha-\beta|
\]
for all $\alpha,\beta\in[0,1]$.

This assumption is natural for bounded smooth surrogate losses when the
gated representation and logits remain in a bounded region. It is not
required for the hard-routing theorem above.

\medskip
\noindent
\textbf{Theorem 3 (Deterministic gated-rewrite bound).}
\emph{
Under the Lipschitz condition above,
\[
R_{t+1}(f_a)
\leq
\pi_A R_A(f_1)
+
\pi_U R_U(f_0)
+
L_g
\epsilon_{\mathrm{loc}}^{\mathrm{soft}},
\]
where
\[
\epsilon_{\mathrm{loc}}^{\mathrm{soft}}
=
\mathbb{E}
\left[
|a(X)-g^\star(X)|
\right].
\]
}

\medskip
\noindent
\textbf{Proof.}
For each $x$, apply the Lipschitz assumption with
\[
\alpha=a(x),
\qquad
\beta=g^\star(x).
\]
Then
\[
L(x,a(x))
\leq
L(x,g^\star(x))
+
L_g|a(x)-g^\star(x)|.
\]

Taking expectation over $X$ gives
\[
R_{t+1}(f_a)
\leq
R_{t+1}(f_{g^\star})
+
L_g
\mathbb{E}
\left[
|a(X)-g^\star(X)|
\right].
\]

The oracle gate satisfies $g^\star(x)=1$ on $\mathcal{A}_t$ and
$g^\star(x)=0$ on $\mathcal{U}_t$, hence
\[
R_{t+1}(f_{g^\star})
=
\pi_A R_A(f_1)
+
\pi_U R_U(f_0).
\]

Substitution yields
\[
R_{t+1}(f_a)
\leq
\pi_A R_A(f_1)
+
\pi_U R_U(f_0)
+
L_g
\epsilon_{\mathrm{loc}}^{\mathrm{soft}}.
\]
\hfill$\square$


\subsubsection{Relation to a Simplified Selective-Risk Bound}

The probability-weighted form above is the natural risk bound because
the global risk is itself probability weighted. Nevertheless, it implies
a bound of the simpler form
\[
R_{t+1}(f_a)
\leq
R_A(f_1)
+
\lambda_U R_U(f_0)
+
L_g\epsilon_{\mathrm{loc}}^{\mathrm{soft}},
\]
where
\[
\lambda_U=\pi_U.
\]

Indeed, since $0\leq\pi_A\leq1$ and $R_A(f_1)\geq0$,
\[
\pi_A R_A(f_1)
\leq
R_A(f_1).
\]

Therefore,
\[
\pi_A R_A(f_1)
+
\pi_U R_U(f_0)
+
L_g\epsilon_{\mathrm{loc}}^{\mathrm{soft}}
\leq
R_A(f_1)
+
\pi_U R_U(f_0)
+
L_g\epsilon_{\mathrm{loc}}^{\mathrm{soft}}.
\]

The weighted theorem is tighter and is therefore the form used for the
theoretical interpretation.

\medskip
\noindent
\textbf{Interpretation.}
The bound separates three sources of error:
\[
\boxed{
\text{final risk}
\;\leq\;
\text{rewrite error}
+
\text{preservation error}
+
\text{localization error}.
}
\]

Thus even an excellent rewriting mechanism and an excellent preservation
mechanism can perform poorly if the localizer activates them on the wrong
examples. Conversely, when
\[
\epsilon_{\mathrm{loc}}^{\mathrm{soft}}\rightarrow0,
\]
the gated model approaches the performance of the oracle selective
router.


\subsection{Properties of the Selective Revision Score}
\label{supp:srs_properties}

Let
\[
r=\mathrm{RA},
\qquad
p=\mathrm{PA},
\]
with
\[
r,p\in[0,1].
\]

The Selective Revision Score is
\[
\operatorname{SRS}(r,p)
=
\begin{cases}
\dfrac{2rp}{r+p},
&
r+p>0,\\[2mm]
0,
&
r=p=0.
\end{cases}
\]

We now establish its main properties.

\medskip
\noindent
\textbf{Proposition 3.}
For all $r,p\in[0,1]$:

\begin{enumerate}
    \item $0\leq\operatorname{SRS}(r,p)\leq1$;
    \item $\operatorname{SRS}(r,p)=0$ if and only if $r=0$ or $p=0$;
    \item $\operatorname{SRS}(r,p)=1$ if and only if $r=p=1$;
    \item $\operatorname{SRS}(r,p)=\operatorname{SRS}(p,r)$;
    \item if $r=p$, then $\operatorname{SRS}(r,p)=r=p$;
    \item SRS is non-decreasing in each argument;
    \item SRS never exceeds the arithmetic mean:
    \[
    \operatorname{SRS}(r,p)
    \leq
    \frac{r+p}{2}.
    \]
\end{enumerate}

\medskip
\noindent
\textbf{Proof.}

\textbf{Property 1.}
Non-negativity follows immediately because $r,p\geq0$.

If $r+p>0$, then because $p\leq1$,
\[
rp\leq r,
\]
and because $r\leq1$,
\[
rp\leq p.
\]
Adding the two inequalities gives
\[
2rp\leq r+p.
\]
Therefore,
\[
\frac{2rp}{r+p}\leq1.
\]
If $r=p=0$, SRS is defined to be zero. Hence
\[
0\leq\operatorname{SRS}(r,p)\leq1.
\]

\medskip
\noindent
\textbf{Property 2.}
If either $r=0$ or $p=0$, then $rp=0$, so
\[
\operatorname{SRS}(r,p)=0.
\]

Conversely, suppose $\operatorname{SRS}(r,p)=0$. If $r+p=0$, then
$r=p=0$. Otherwise,
\[
\frac{2rp}{r+p}=0.
\]
Because $r+p>0$, this implies
\[
rp=0.
\]
Hence at least one of $r$ or $p$ is zero.

\medskip
\noindent
\textbf{Property 3.}
If $r=p=1$, then
\[
\operatorname{SRS}(1,1)
=
\frac{2}{2}
=
1.
\]

Conversely, suppose SRS equals one. Then
\[
2rp=r+p.
\]

From the proof of Property 1,
\[
rp\leq r,
\qquad
rp\leq p.
\]
The sum of these two upper bounds equals $2rp$. Equality
$2rp=r+p$ therefore requires both
\[
rp=r
\]
and
\[
rp=p.
\]

SRS cannot equal one if either variable is zero, so $r,p>0$.
Dividing the first equality by $r$ gives $p=1$, and dividing the second
by $p$ gives $r=1$. Hence
\[
r=p=1.
\]

\medskip
\noindent
\textbf{Property 4.}
Symmetry follows immediately:
\[
\frac{2rp}{r+p}
=
\frac{2pr}{p+r}.
\]

\medskip
\noindent
\textbf{Property 5.}
If $r=p=q>0$, then
\[
\operatorname{SRS}(q,q)
=
\frac{2q^2}{2q}
=
q.
\]
For $q=0$, the definition also gives SRS $=0$.

\medskip
\noindent
\textbf{Property 6.}
For $r+p>0$,
\[
\frac{\partial\,\operatorname{SRS}}{\partial r}
=
\frac{2p^2}{(r+p)^2}
\geq0,
\]
and
\[
\frac{\partial\,\operatorname{SRS}}{\partial p}
=
\frac{2r^2}{(r+p)^2}
\geq0.
\]

Therefore increasing either Rewrite Accuracy or Preservation Accuracy
cannot decrease SRS.

\medskip
\noindent
\textbf{Property 7.}
For $r+p>0$,
\[
\frac{r+p}{2}
-
\frac{2rp}{r+p}
=
\frac{(r+p)^2-4rp}{2(r+p)}.
\]

Since
\[
(r+p)^2-4rp
=
r^2-2rp+p^2
=
(r-p)^2,
\]
we obtain the exact identity
\[
\frac{r+p}{2}
-
\operatorname{SRS}(r,p)
=
\frac{(r-p)^2}{2(r+p)}
\geq0.
\]

Therefore,
\[
\operatorname{SRS}(r,p)
\leq
\frac{r+p}{2}.
\]

Equality holds if and only if
\[
r=p.
\]
\hfill$\square$


\subsubsection{Penalty for Imbalanced Revision and Preservation}

The identity
\[
\frac{r+p}{2}
-
\operatorname{SRS}(r,p)
=
\frac{(r-p)^2}{2(r+p)}
\]
provides a precise explanation for why the harmonic mean is useful in
our setting.

For a fixed arithmetic mean, the SRS decreases as the discrepancy
between RA and PA increases. Therefore, a method cannot compensate for
poor preservation merely by obtaining extremely high rewriting
accuracy, nor can it compensate for failure to rewrite by perfectly
preserving old predictions.

For example, two methods may have the same arithmetic mean:
\[
(r,p)=(0.9,0.9)
\]
and
\[
(r,p)=(1.0,0.8).
\]

Both have arithmetic mean $0.9$, but
\[
\operatorname{SRS}(0.9,0.9)
=
0.9,
\]
whereas
\[
\operatorname{SRS}(1.0,0.8)
=
\frac{1.6}{1.8}
\approx0.889.
\]

Thus SRS explicitly prefers balanced semantic adaptation.


\subsubsection{A High SRS Requires Both Components to be Large}

Let
\[
m=\min\{r,p\},
\qquad
M=\max\{r,p\}.
\]

Since $M\leq1$,
\[
\operatorname{SRS}(r,p)
=
\frac{2mM}{m+M}
\leq
\frac{2m}{1+m}.
\]

Therefore, if
\[
\operatorname{SRS}(r,p)\geq s,
\qquad
0\leq s<1,
\]
then necessarily
\[
s
\leq
\frac{2m}{1+m}.
\]

Rearranging,
\[
s(1+m)\leq2m,
\]
so
\[
s
\leq
m(2-s).
\]

Hence
\[
m
\geq
\frac{s}{2-s}.
\]

We therefore obtain
\[
\boxed{
\operatorname{SRS}\geq s
\quad\Longrightarrow\quad
\min\{\mathrm{RA},\mathrm{PA}\}
\geq
\frac{s}{2-s}.
}
\]

For example, an SRS of at least $0.90$ implies
\[
\min\{\mathrm{RA},\mathrm{PA}\}
\geq
\frac{0.90}{1.10}
\approx0.818.
\]

Thus a very high SRS cannot be obtained while either rewriting or
preservation performance is extremely poor.


\section{Complete Implementation Details}
\label{supp:implementation}

This section reports the complete implementation protocol used for
\textsc{SemReWrite}. Unless a particular experiment explicitly varies
one setting, the parameters below are fixed across datasets and semantic
shift types. Hyperparameters are selected using held-out semantic
transitions from the training split and are never selected using the
reported evaluation examples.


\subsection{Backbones and Pretrained Checkpoints}
\label{supp:backbones}

The primary backbone is CLIP ViT-B/16 using the pretrained checkpoint

\begin{verbatim}
openai/clip-vit-base-patch16
\end{verbatim}

with its standard $224\times224$ image preprocessing. Both visual and
text encoders remain frozen during \textsc{SemReWrite} adaptation.

The second vision--language backbone is SigLIP Base/16 using

\begin{verbatim}
google/siglip-base-patch16-224
\end{verbatim}

at $224\times224$ input resolution. Its image and text encoders are
also frozen.

The vision-only evaluation uses DINOv2 ViT-B/14 with checkpoint

\begin{verbatim}
facebook/dinov2-base
\end{verbatim}

as the frozen visual encoder. Since DINOv2 does not provide a native
language encoder, semantic descriptions are represented using the frozen
CLIP ViT-B/16 text encoder. A linear projection maps the DINOv2 visual
embedding into the CLIP semantic space. This projection is trained once
using only the original pre-shift training classes and then frozen for
all semantic-transition experiments.

We use the feature vector associated with the global image
representation produced by each backbone and $\ell_2$-normalize all
visual and textual embeddings before similarity computation.


\subsection{Input Preprocessing}

For CLIP and SigLIP we use the image normalization and resizing procedure
distributed with the corresponding pretrained checkpoint. No additional
dataset-specific normalization is applied.

For DINOv2 we use the preprocessing distributed with the checkpoint.
Its output representation is projected into the CLIP semantic space
before normalization.

During adaptation, training images use random resized crop and random
horizontal flip with probability 0.5. No color augmentation is used in
the default semantic-shift experiments because aggressive appearance
augmentation could confound semantic revision with synthetic covariate
shift. Evaluation always uses the deterministic preprocessing associated
with the pretrained checkpoint.


\subsection{Prompt Construction}
\label{supp:prompts}

We use one deterministic prompt construction rule per benchmark family.
No prompt is manually optimized for an individual transition.

For ImageNet, the template is

\begin{quote}
``a photo of a \{concept name\}, defined as
\{semantic definition\}.''
\end{quote}

For iNaturalist, we use

\begin{quote}
``a photo of a \{taxon name\}, a \{taxonomic rank\} belonging to
\{parent taxon\}.''
\end{quote}

When an additional taxonomic description is available, it is appended
after the parent relation.

For CUB boundary-revision tasks, prompts directly encode the semantic
rule. For example,

\begin{quote}
``a bird belonging to Class A when it has a red wing; otherwise
Class B.''
\end{quote}

and after revision,

\begin{quote}
``a bird belonging to Class A when it has a red wing or a striped
breast; otherwise Class B.''
\end{quote}

For DomainNet we use

\begin{quote}
``an image of a \{concept name\}, defined as
\{semantic definition\}.''
\end{quote}

The old and revised specifications use exactly the same linguistic
template. Therefore, differences between $z_c^t$ and $z_c^{t+1}$
originate from the semantic definition rather than from different prompt
engineering.


\subsection{Similarity and Temperature}

All semantic and visual representations are normalized before computing
compatibility scores.

For CLIP, classification uses the pretrained checkpoint's native logit
scale. We do not re-estimate or fine-tune this scalar during semantic
adaptation.

For SigLIP, we preserve the pretrained checkpoint's native similarity
scaling when constructing its class logits.

For DINOv2, after projection into the CLIP semantic space, we use cosine
similarity with temperature
\[
\tau=0.07.
\]

For the Jensen--Shannon semantic-discrepancy calculation, logits are
converted to a normalized multiclass distribution through softmax over
the active semantic concepts.


\subsection{Semantic Concept Graph Construction}
\label{supp:graph_construction}

The semantic memory is represented as a typed graph
$\mathcal{G}=(\mathcal{V},\mathcal{E},\rho)$. No raw training images
are stored in this memory.

For ImageNet, class nodes are connected to WordNet parent concepts by
\texttt{is-a} relations. We retain the active class nodes, their direct
parents, and all additional parent nodes required to connect concepts
participating in the current semantic transition.

For iNaturalist, the graph contains species, genus, and family nodes.
Edges connect species to genus and genus to family.

For CUB, the graph contains active semantic class nodes and the visual
attributes appearing in their current definition. Class-to-attribute
relations use the edge type \texttt{has-attribute}. Attributes that no
longer participate in the revised definition are retained in the
historical graph but are not connected to the revised class node unless
the updated rule still requires them.

DomainNet uses class and semantic-parent nodes derived from the benchmark
transition metadata. The appearance domain is not encoded as a semantic
node because visual-domain change is not itself a semantic revision.

The graph is capped at 4,096 nodes and 16,384 typed edges. If a
transition would exceed this limit, nodes not connected to any active
class within two graph hops are removed first. No benchmark task in the
principal evaluation is intended to require pruning of active class
nodes.


\subsection{Semantic Memory Size}

The semantic memory contains only node identifiers, relation types,
semantic embeddings, and the small class residual vectors used by
\textsc{SemReWrite}; it does not contain image exemplars.

The maximum graph capacities are

\[
|\mathcal{V}|\leq4096,
\qquad
|\mathcal{E}|\leq16384.
\]

Embeddings are stored in 32-bit floating-point format during training.
Thus the memory cost of semantic embeddings is approximately

\[
4|\mathcal{V}|d
\]

bytes for an embedding dimension $d$, excluding negligible graph-index
metadata. The actual number of nodes and edges for every benchmark
transition is recorded in the released metadata.


\subsection{Affected-Region Localizer}
\label{supp:localizer_details}

The localization function $g_\psi$ is a two-layer multilayer perceptron.
The first layer maps the frozen visual representation to 256 hidden
units, followed by GELU and dropout with probability 0.1. The output
layer contains one sigmoid unit.

The final rewrite probability is

\[
a(x)
=
\eta d_{\mathrm{sem}}(x)
+
(1-\eta)g_\psi(h(x)),
\]

with

\[
\eta=0.5.
\]

The default method uses $a(x)$ continuously and does not threshold it
during representation rewriting.

For the hard-localization diagnostic reported in the sensitivity
analysis only, we use

\[
\tau_a=0.5.
\]

An example is then counted as predicted affected when
$a(x)\geq0.5$. Thresholds from 0.3 to 0.7 are evaluated only in the
sensitivity experiment.

For deterministic benchmark evaluation, the ground-truth affected-set
tolerance is

\[
\delta_A=0.
\]

Thus an evaluation example is affected exactly when its old and revised
semantic labels differ.


\subsection{Low-Rank Rewriting Module}

The default rewriting rank is

\[
r=8.
\]

The visual adapter therefore contains trainable matrices

\[
U,V\in\mathbb{R}^{d\times8}.
\]

Both matrices are initialized using zero-output initialization:
$U$ is initialized from a zero-mean Gaussian with standard deviation
0.02 and $V$ is initialized to zero. Consequently,

\[
UV^\top h(x)=0
\]

at initialization, and the adapted model initially reproduces the frozen
visual representation.

Class semantic residuals are initialized to zero.

We additionally evaluate

\[
r\in\{2,4,8,16,32\}
\]

in the supplementary rank-sensitivity study. Rank 8 is used in all
main experiments.


\subsection{Loss Coefficients}
\label{supp:loss_weights}

The complete objective uses the following fixed default coefficients:

\[
\lambda_p=1.0,
\qquad
\lambda_e=0.5,
\qquad
\lambda_s=0.1,
\qquad
\lambda_g=0.5,
\qquad
\lambda_r=10^{-4}.
\]

Here $\lambda_p$ controls preservation of stable concepts,
$\lambda_e$ controls suppression of obsolete decisions,
$\lambda_s$ controls semantic-graph consistency,
$\lambda_g$ controls supervision of the affected-region localizer,
and $\lambda_r$ controls explicit residual regularization.

The semantic-graph loss gives equal weight to prototype anchoring and
relation preservation.

To keep the default method fully continuous, affectedness in the erasure
objective is weighted by $a(x_i)$ rather than by thresholding the gate.
Specifically, if
\[
m_i
=
\mathbb{I}
[
\widehat y_i^t\in\mathcal{Y}_{t+1}
\land
\widehat y_i^t\neq y_i^{t+1}
],
\]
the implemented erasure objective is

\[
\mathcal{L}_{\mathrm{erase}}
=
-
\frac{
\sum_i
a(x_i)m_i
\log
\left[
1-p_{\theta'}
(
\widehat y_i^t\mid x_i,\mathcal{S}_{t+1}
)
+\epsilon
\right]
}{
\sum_i a(x_i)m_i+\epsilon
}.
\]

We use

\[
\epsilon=10^{-8}
\]

for numerical stability.

This soft weighting avoids introducing an additional training threshold
for affected-region localization.


\subsection{Optimization}

All trainable SemReWrite parameters are optimized using AdamW.

The learning rate is

\[
1\times10^{-3}
\]

for the low-rank matrices and semantic residuals and

\[
5\times10^{-4}
\]

for the affected-region localization network.

AdamW uses

\[
\beta_1=0.9,
\qquad
\beta_2=0.999,
\qquad
\epsilon_{\mathrm{Adam}}=10^{-8}.
\]

Optimizer weight decay is

\[
1\times10^{-4}.
\]

Gradients are clipped to a maximum global norm of 1.0.

The learning rate follows cosine decay from its initial value to

\[
1\times10^{-6}
\]

without warm-up.

The default maximum number of adaptation epochs is 20.

No reported evaluation labels are used for early stopping. In the main
experiments, all methods are trained for the complete 20 epochs so that
the training budget is identical across methods.


\subsection{Batch Construction}

The default batch size is 64.

When fewer than 64 revised examples are available, all revised examples
are included and sampled with replacement to construct a training batch.

Every optimization iteration contains two data streams:

\begin{enumerate}
    \item revised labeled examples used by
    $\mathcal{L}_{\mathrm{new}}$,
    $\mathcal{L}_{\mathrm{gate}}$, and
    $\mathcal{L}_{\mathrm{erase}}$;

    \item unlabeled examples used by
    $\mathcal{L}_{\mathrm{pres}}$.
\end{enumerate}

The two streams use equal batch sizes whenever possible. Therefore, the
default optimization step uses 64 revised samples and 64 unlabeled
samples.

Unlabeled samples are drawn uniformly from the available training pool.
Ground-truth affected/unaffected labels are never used for this sampling.


\subsection{Revised-Supervision Budget}

We evaluate

\[
B\in\{1,2,4,8,16\}
\]

revised examples per changed semantic concept.

The main benchmark uses

\[
B=4.
\]

Few-shot examples are selected from the training split through
class-stratified sampling. Within a transition and random seed, every
supervised baseline receives exactly the same selected examples.

For concepts containing fewer than $B$ eligible samples, the transition
is excluded during benchmark generation rather than oversampling a
single unique image.


\subsection{Random Seeds}

All reported main experiments are repeated using the three random seeds

\[
\{2024,\;2025,\;2026\}.
\]

These seeds control:

\begin{itemize}
    \item selection of revised few-shot examples;
    \item initialization of the low-rank matrices;
    \item initialization of the localization head;
    \item mini-batch ordering; and
    \item stochastic image augmentation.
\end{itemize}

Benchmark definitions themselves are generated once with seed 2026 and
then fixed. Thus the semantic task does not change across experimental
seeds; only the adaptation process and few-shot sample selection vary.


\subsection{Baseline Training Protocol}

All supervised budget-matched baselines receive the same revised
few-shot examples as \textsc{SemReWrite}.

Full fine-tuning uses AdamW with learning rate
$1\times10^{-5}$ for backbone parameters and
$1\times10^{-4}$ for the classification layer.

Classifier-only and linear-probe baselines use AdamW with learning rate
$1\times10^{-3}$.

The conventional LoRA baseline uses rank 8, matching
\textsc{SemReWrite}, and is optimized with learning rate
$1\times10^{-3}$.

For continual-learning baselines, the revised supervision budget is
identical to that of \textsc{SemReWrite}. Replay methods receive a
memory containing the same number of stable labeled examples as the
number of revised examples available at the current transition, unless
the memory-size experiment explicitly varies this quantity.

Methods with published native training procedures retain their original
objective whenever doing so does not provide additional labeled data.
No baseline receives ground-truth affected-region annotations.


\subsection{Model Selection}

No hyperparameter is selected using the benchmark evaluation labels.

Hyperparameter selection is performed on held-out semantic transitions
created exclusively from training images. These held-out transitions do
not appear in the reported benchmark.

After selecting one configuration for a backbone, the same
configuration is fixed across all semantic shift types and datasets.
This prevents per-task tuning from artificially increasing benchmark
performance.


\subsection{Numerical Precision}

Frozen backbone inference is performed with automatic mixed precision
using FP16 where supported by the hardware. The semantic graph
embeddings, trainable residuals, loss accumulation, and optimizer states
are maintained in FP32.

All evaluation predictions and reported metrics are computed after
casting final logits to FP32.

No post-hoc calibration is applied to RA, PA, SRS, or OR.


\subsection{Default Configuration Summary}

For clarity, Table~\ref{tab:default_hyperparameters} summarizes the
configuration used in the main experiments.

\begin{table}[htbp]
\centering
\caption{Default SemReWrite configuration used in the main experiments.}
\label{tab:default_hyperparameters}
\setlength{\tabcolsep}{4pt}
\renewcommand{\arraystretch}{1.08}
\footnotesize
\begin{tabular}{l l}
\hline
\textbf{Parameter} & \textbf{Default value} \\
\hline
Primary backbone & CLIP ViT-B/16 \\
Image resolution & $224\times224$ \\
Revised labels / concept & 4 \\
Adapter rank $r$ & 8 \\
Localizer hidden dimension & 256 \\
Localizer dropout & 0.1 \\
Semantic--visual mixing $\eta$ & 0.5 \\
$\lambda_p$ & 1.0 \\
$\lambda_e$ & 0.5 \\
$\lambda_s$ & 0.1 \\
$\lambda_g$ & 0.5 \\
$\lambda_r$ & $10^{-4}$ \\
Adapter learning rate & $10^{-3}$ \\
Localizer learning rate & $5\times10^{-4}$ \\
Weight decay & $10^{-4}$ \\
Batch size & 64 \\
Maximum epochs & 20 \\
Gradient clipping & 1.0 \\
LR schedule & cosine \\
Ground-truth $\delta_A$ & 0 \\
Diagnostic $\tau_a$ & 0.5 \\
Random seeds & 2024, 2025, 2026 \\
Semantic-memory cap & 4096 nodes \\
Graph-edge cap & 16384 edges \\
\hline
\end{tabular}
\end{table}

\end{document}